%% file: main.tex
\documentclass[10pt,journal,compsoc]{IEEEtran}

\ifCLASSOPTIONcompsoc
  \usepackage[nocompress]{cite}
\else
  \usepackage{cite}
\fi
\usepackage{CJKutf8}
\usepackage[pdftex]{graphicx}
\usepackage{textcomp}
\usepackage{bm}
\usepackage{xcolor}
\usepackage{color} 
\usepackage{amssymb}
\usepackage{amsmath}
\usepackage{multirow} 
\usepackage{booktabs} 
\usepackage{float}
\usepackage[pagebackref=true,breaklinks=true,letterpaper=true,colorlinks,citecolor=blue,linkcolor=blue,bookmarks=false]{hyperref}
\ifCLASSINFOpdf
\else
\fi
\usepackage{comment}
\begin{document}
% \title{Lodge++: Beat-aligned Long Dance Generation with Genre Consistency}
\title{Lodge++: High-quality and Long Dance Generation with Vivid Choreography Patterns}
% \title{Lodge++: High-quality and Long-duration Dance Generation with Robust Choreography}
% \title{Lodge++: Coarse-to-Fine 3D Long Dance Generation Network}

% \IEEEcompsocitemizethanks is a special \thanks that produces the bulleted
% one-size-fits-all approach a daunting prospect. For instance, compsoc journal papers have the author affiliations above the "Manuscript received ..."  text while in non-compsoc journals this is reversed. Sigh.

\author{Ronghui~Li,%~\IEEEmembership{Member,~IEEE,}
        Hongwen~Zhang,%~\IEEEmembership{Fellow,~OSA,}
        % Hongwen~Zhang,
        Yachao~Zhang,
        Yuxiang~Zhang,\\
        Youliang~Zhang,
        Jie~Guo,
        Yan~Zhang,
        Xiu~Li,
        and Yebin~Liu,~\IEEEmembership{Member,~IEEE,}% <-this % stops a space
\IEEEcompsocitemizethanks{\IEEEcompsocthanksitem Ronghui Li, Yachao Zhang, Youliang Zhang and Xiu Li, are with the Shenzhen International Graduate School, Tsinghua University, Shenzhen 518055, China. 
E-mail: \{lrh22,zhangyou24\}@mails.tsinghua.edu.cn; \{yachaozhang,li.xiu\}@sz.tsinghua.edu.cn
%E-mail:\left\{lrh22,zhangyou24\right\}$\@$mails.tsinghua.edu.cn;li.xiu$\@$sz.tsinghua.edu.cn

\IEEEcompsocthanksitem Yuxiang Zhang and Yebin Liu are with Department of Automation,
Tsinghua University, Beijing 100084, China.
E-mail: yx-z19@mails.tsinghua.edu.cn;liuyebin@mail.tsinghua.edu.cn

%E-mail:yx-z19@mails.tsinghua.edu.cn; liuyebin@mail.tsinghua.edu.cn

\IEEEcompsocthanksitem Jie Guo is with Peng Cheng Laboratory, Shenzhen 518000, China. E-mail: bitguojie@gmail.com

\IEEEcompsocthanksitem Hongwen Zhang is with the School of Artificial Intelligence, Beijing Normal University. Beijing 100875, China. E-mail: zhanghongwen@bnu.edu.cn

\IEEEcompsocthanksitem Yan Zhang is with the  Meshcapade, Tübingen, 72072, Germany. E-mail: yan@meshcapade.com

Corresponding authors: Xiu Li; Yebin Liu.
}
% Y
% GA, 30332.\protect\\
% %note need leading \protect in front of \\ to get a newline within \thanks as
% E-mail: see http://www.michaelshell.org/contact.html
% \IEEEcompsocthanksitem J. Doe and J. Doe are with Anonymous University.}% <-this % stops a space
% \thanks{Manuscript received August 14, 2024.}
}
\newcommand{\yan}[1]{{\color{blue}[\bf \em Yan: {#1}]}}
\newcommand{\lrh}[1]{{\color{red}[\bf \em lrh: {#1}]}}

\input{sec/0_Abstract}

\maketitle

% is not selected <OR> if conference mode is selected - because compsoc
% conference papers position the abstract like regular (non-compsoc)
\IEEEdisplaynontitleabstractindextext

% \fi
\IEEEpeerreviewmaketitle

% \ifCLASSOPTIONcompsoc
% \IEEEraisesectionheading{\section{Introduction}\label{sec:introduction}}
% \else
% \section{Introduction}
% \label{sec:introduction}
% \fi

\input{sec/1_Introduction}

\input{sec/2_RelatedWork}
\input{sec/3_Methodology}
\input{sec/4_Experiments}

\input{sec/5_Conclusion}

\input{sec/6_Appendix}

\ifCLASSOPTIONcaptionsoff
  \newpage
\fi
% \newpage
\bibliographystyle{IEEEtran}
\bibliography{IEEEabrv,bibtex}

\end{document}

%% file: sec/0_Abstract.tex
\IEEEtitleabstractindextext{%
\begin{abstract}
We propose Lodge++, a choreography framework to generate high-quality, ultra-long, and vivid dances given the music and desired genre. To handle the challenges in computational efficiency, the learning of complex and vivid global choreography patterns, and the physical quality of local dance movements, Lodge++ adopts a two-stage strategy to produce dances from coarse to fine.
In the first stage, a global choreography network is designed to generate coarse-grained dance primitives that capture complex global choreography patterns.
In the second stage, guided by these dance primitives, a primitive-based dance diffusion model is proposed to further generate high-quality, long-sequence dances in parallel, faithfully adhering to the complex choreography patterns.
Additionally, to improve the physical plausibility, Lodge++ employs a penetration guidance module to resolve character self-penetration, a foot refinement module to optimize foot-ground contact, and a multi-genre discriminator to maintain genre consistency throughout the dance.
Lodge++ is validated by extensive experiments, which show that our method can rapidly generate ultra-long dances suitable for various dance genres, ensuring well-organized global choreography patterns and high-quality local motion. The project page is \href{https://li-ronghui.github.io/lodgepp}{https://li-ronghui.github.io/lodgepp}.
% Extensive quantitative and qualitative experiments demonstrate the efficacy of our method.
\end{abstract}

\begin{IEEEkeywords}
3D motion generation, digital human, generative model, diffusion, animation.
\end{IEEEkeywords}}

%% file: sec/1_Introduction.tex
\ifCLASSOPTIONcompsoc
\IEEEraisesectionheading{\section{Introduction}\label{sec:introduction}}
\else
\section{Introduction}
\label{sec:introduction}
\fi
% \begin{CJK}{UTF8}{gbsn}
% {\color{red}感谢大家指导本文的writting，这个版本的Introduction已根据建议修改，Related Work和Method正在修改中。}
% \end{CJK}
\IEEEPARstart{M}{usic} and dance have been fundamental aspects of human culture for centuries. In the modern age of digital content, there is a high demand for 3D dance content in industries such as film, animation, virtual reality, and social media. However, traditional methods of obtaining 3D dance sequences require professional dancers and motion capture equipment, leading to high costs and low efficiency that can not meet the industry's growing demands. Consequently, automatically generating dance sequences based on music has become an important research topic, which not only helps the film and television industry quickly produce 3D dance assets but also assists dancers in their choreography creation.

In recent years, the music-driven dance generation has made significant progress with the development of generative models~\cite{goodfellow2014generative,radford2015unsupervised,bengio2013representation,diffusionbeatgan,DDPM,DDIM}. However, most efforts~\cite{aist++,sun2022you,kim2022brand,edge,li2022danceformer} have focused on producing high-quality dance sequences that last only a few seconds, leaving efficient methods for generating longer sequences largely unexplored. This limitation hampers the ability to quickly create high-quality, minute-long dance sequences, reducing their practical applicability.

In this paper, we identify the following challenges towards the goal of high-quality and long-sequence dance generation with appropriate choreography patterns:
% our goal is to effectively generate high-quality long-sequence dances with appropriate global choreography patterns, which presents the following challenges:
% Generating high-quality long sequences of dance poses several challenges:
\textbf{1) Computational Efficiency:} Generating long sequences significantly increases computational demands. Reducing the algorithm's computational load and improving training and inference efficiency are primary concerns.
\textbf{2) Global Choreography Patterns:} 
\textit{Long sequence dance contains complex global choreographic patterns. }
Dancers enhance the expressiveness and memorable moments of the dance by performing repeated or mirrored key movements. They also express heightened or shifting emotions through progressive or turning actions.
Meaningless, chaotic movements or excessive repetition should be avoided. 
% Additionally, the global choreography patterns of the dance needs to align with the music. 
The challenge lies in the fact that different dance genres have unique choreographic patterns, which are complex and difficult to summarize with simple rules.
\textbf{3) Local Dance Quality:} 
Achieving graceful and expressive dance movements while maintaining physical realism is also a significant challenge.
Meanwhile, the rhythm of the dance should align with the musical beats as closely as possible.

\begin{figure}[t]
\begin{center}
\includegraphics[width=0.45\textwidth]{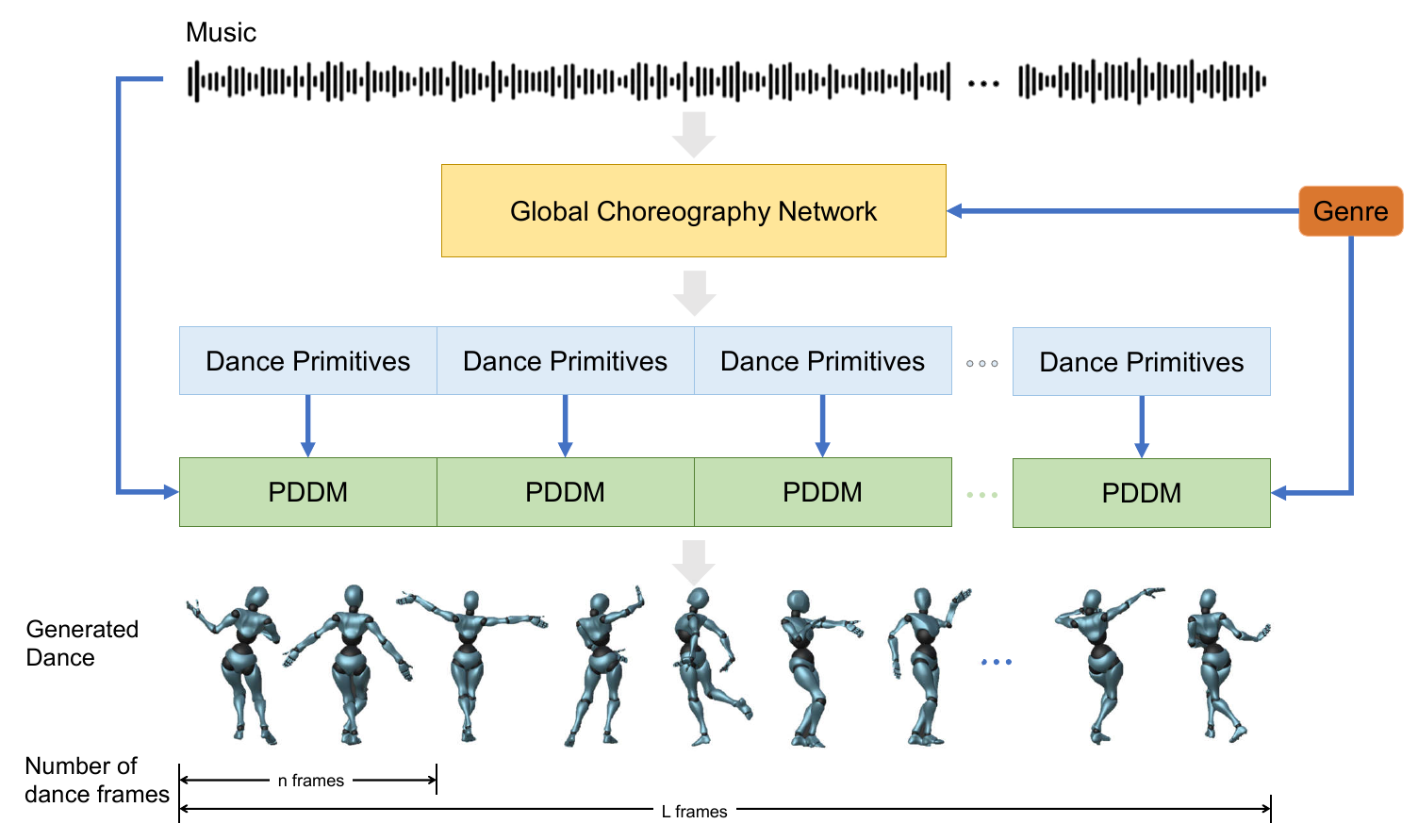}
% \vspace{-3mm}
\caption{An overview of Lodge++. ``PDDM" is the proposed Primitive-based Dance Diffusion Model.
Given the extremely long music and desired dance genre as input, Lodge++ uses the Global Choreography Network to generate dance primitives that contain global choreography patterns. Then, it leverages the parallel PDDM network to generate high-quality and coherent long-sequence choreographic dance.}	
% \vspace{-6mm}
\label{fig:framework}
\end{center}
\end{figure}

Previous works have made various efforts to address these challenges. To reduce the computational cost of the training process, many have adopted autoregressive architectures~\cite{aist++,kim2022brand,edge}, training music-to-dance generation networks within small time windows, typically 2-5 seconds. During the generation phase, long dance sequences are produced autoregressively through a sliding window. However, these approaches fail to learn global choreography patterns because they focus on short-term dependencies. The autoregressive nature of these methods accumulates prediction errors, resulting in motion freezing or meaningless swaying~\cite{yang2023longdancediff,bailando++}.
To model global choreography patterns while reducing computational load, some methods adopt a two-stage architecture. The first stage compresses dance sequences into a compact latent space, and the second stage learns the overall coarse-grained music-to-dance dependencies in that latent space. Bailando~\cite{bailando} uses a VQ-VAE for compression and then employs a GPT sequence model to learn the music-dance pairing. However, due to the limitations of VQ-VAE's representational capacity, the generated dances still suffer from issues such as foot-ground contact problems and physical unrealism, including penetration and unnatural transitional motions.

% Previous works have made various efforts. To reduce the computational cost of the training process, many works have adopted autoregressive architectures~\cite{aist++,kim2022brand,edge}, training a music-to-dance generation network within a small time window, typically 2-5 seconds. 
% During the generation phase, long sequences of dance are generated autoregressively through a sliding window. However, this approach fails to learn global choreography patterns. The autoregressive nature of these methods accumulate prediction errors, resulting in motion freezing or meaningless swaying~\cite{yang2023longdancediff,bailando++}.
% To further model global choreography patterns while reducing computational load, some methods adopt a two-stage architecture. The first stage compresses dance sequences into a compact latent space, and the second stage learns the overall coarse-grained music-to-dance dependencies at the latent space. Bailando~\cite{bailando} maintains a compressed latent space using VQ-VAE and then employs a GPT sequence model to learn the music-dance pairing. However, due to the limitations of VQ-VAE’s representational capacity, the generated dances still suffer from issues like foot-ground contact problems and physical unrealism, such as penetration and unnatural transitional motion.

In summary, existing methods do not effectively address the challenges of computational efficiency, global choreography patterns, and local dance quality.
They treat music-driven dance generation as a sequence-to-sequence translation task, which may not fully capture the complexities of dance creation. According to~\cite{dance_design,dance_design2,blom1982intimate,choreomaster}, dance is typically choreographed in a coarse-to-fine manner: the global choreography is first established, followed by the refinement of detailed local movements. 
Inspired by this, we design Lodge++, a two-stage architecture that includes a computation-friendly Global Choreography Network and a local Primitive-based Dance Diffusion Model, with dance primitives serving as an intermediate representation to transfer global choreography patterns across two stages.

The Global Choreography Network effectively models complex and vivid global choreography patterns in a compressed VQ-VAE latent space and generates dance primitives.
These dance primitives consist of expressive 8-frame key motions of the body’s main limbs with high kinetic energy.
The dance primitives are designed inspired by Labanotation~\cite{guest2013labanotation}, a system widely used for recording various dance genres through key movements of the body’s main limbs.
The proposed dance primitives have the following main advantages: i) they have a flexible quantity, making them suitable for various dance genres;
ii) they contain rich semantic information, allowing them to convey complex choreographic patterns;
iii) they have strong expressive characteristics, enabling the subsequent diffusion model to generate more dynamic movements and prevent monotony.

Based on the above design, the role of the Primitive-based Dance Diffusion Model (PDDM) is to generate high-quality long-sequence dances in parallel under the guidance of dance primitives.
% To enhance PDDM's ability to generate high-quality dance movements, 
To further improve the visual quality of dance movements, 
we design a Foot Refine Block module and multi-genre discriminator to enhance foot-ground contact quality while maintaining genre consistency. Moreover, we propose a novel Signed Distance Field (SDF)-based Penetration Guidance strategy to address the self-penetration issue in complex dance movements, improving the physical plausibility of the generated dances.

% Guided by dance primitives, PDDM can parallelly generate long-sequence dances with both vivid global choreography patterns and graceful local movements.

A preliminary version of this work is Lodge~\cite{li2024lodge}, which was published in CVPR 2024. Lodge proposes a two-stage coarse-to-fine choreography framework that uses Global Diffusion to model choreography patterns and Local Diffusion to generate high-quality dance movements in parallel.
The Global Diffusion generates ``characteristic dance primitives", which are sparse but expressive key movements.
These characteristic dance primitives are further mirrored and shifted to align with the musical beats according to basic choreographic rules~\cite{choreomaster,li2024lodge}, guiding the Local Diffusion to generate high-quality long-sequence dances that conform to choreographic rules. However, Lodge relies heavily on predefined choreographic rules, which are overly simplistic and not entirely applicable to all dance genres. This limitation prevents Lodge from learning complex and vivid choreographic patterns. Furthermore, transmitting characteristic dance primitives to Local Diffusion through diffusion guidance introduces discrepancies between training and inference, leading to unnatural movements such as sudden changes. Additionally, the generated movements sometimes exhibit self-penetration issues.

We extend the preliminary version~\cite{li2024lodge} in the following ways. First, we replace Lodge’s Global Diffusion with a VQ-VAE+GPT-based Global Choreography Network and substitute the fixed-number characteristic dance primitives with flexible-number coarse-grained dance primitives. This enables learning more complex and vivid global choreography patterns suitable for any dance genre. Second, we upgrade Lodge’s Local Diffusion to a Primitive-based Dance Diffusion Model, reducing the train-test gap and solving the issue of abrupt motion transitions caused by using dance primitives and diffusion guidance in Local Diffusion. Third, Lodge++ introduces an SDF-based penetration guidance strategy to address the self-penetration issue in complex dance movements.

In summary, the main contributions of Lodge++ can be summarized as follows:
% \yan{here you might need to explicitly say that this work is extended from Lodge. Compared to Lodge, what are the additional novelties? Extending CVPR to PAMI normally needs 30\% new contributions.}
% \lrh{Thanks for your suggestion! I will list the main differences with Lodge at a new paragraph.}

\begin{itemize}
\item We propose Lodge++, a two-stage long-sequence dance generation network composed of a computation-friendly Global Choreography Network and a Primitive-based Dance Diffusion Model. We design dance primitives as an intermediate-level representation that not only allows for the efficient transmission of global choreography patterns but also boosts the expressiveness of the generated dance.
% By leveraging these dance primitives, Lodge++ effectively models complex choreography and enables the creation of extremely long, high-quality, and physically plausible dance movements in seconds.
\item We design a new global choreography network using VQ-VAE+GPT to effectively model complex and vivid global choreography patterns that are suitable for any dance genre.
\item We develop the Primitive-based Dance Diffusion Model capable of parallel generation of long-sequence dances driven by a flexible number of dance primitives.
\item 
We propose a novel, training-free penetration guidance strategy that utilizes the signed distance field to shift denoising process, thereby effectively addressing character self-penetration issues.
We also introduce a Foot Refine Block to optimize foot-ground contact and a Multi-Genre Discriminator to maintain genre consistency.
\end{itemize}

%% file: sec/2_RelatedWork.tex
\section{Related Work}
\label{sec:related}

\subsection{Human Motion Synthesis}
Human motion synthesis has been a prominent area of research in computer vision and computer graphics.

% \yan{motion graph methods, list some works, state their pros and cons...}

% \yan{motion matching methods, list some works, state their pros and cons...}

% \yan{data-driven and learning-based methods, such as VAE, diffusion, etc., list some works, state their pros and cons...}

% \yan{physical simulation-based methods, list some works, state their pros and cons...}

% \yan{why our method uses diffusion models instead of others? give the reason...}

% Early works \yan{people are still working on motion graphs, and they are not early and out-of-date..dont trigger them.. }\lrh{haha,Yes.We can't ignore them} 
% primarily utilized motion-graph based algorithms to synthesize human motion~\cite{kovar2008motion,arikan2002interactive,geofeature}, which essentially involved cutting and stitching existing motion datasets according to task-specific rules. Although these methods can produce high-quality motion, they require maintaining complex graph structures and often lack diversity in the synthesized motions.

\textbf{Motion graph}~\cite{kovar2008motion, arikan2002interactive, mg1, mg2, geofeature} have been extensively employed for human motion synthesis for decades. These methods construct a graph where nodes represent motion clips from a dataset, and edges represent possible transitions between motions. New motions are synthesized by traversing the graph according to specific rules or constraints. While motion graphs can produce high-quality and realistic motions by utilizing real motion capture data, they require maintaining complex graph structures and substantial preprocessing. Moreover, the diversity and flexibility of the synthesized motions are limited to combinations of existing motion clips, making it challenging to generate novel or highly varied motions.

Similarly, \textbf{motion matching} techniques~\cite{kim2009synchronized, holden2016deep,holden2017phase,holden2020learned,starke2020local,starke2021neural,deepphase} aim to match user inputs or constraints to the most suitable motion frames from a database. These methods are efficient and responsive, making them suitable for interactive applications like video games. However, their performance heavily depends on the richness of the motion database. They struggle to generate motions that are not present in the dataset and lack the capability to produce long, complex dance sequences automatically.

In recent years, with the development of deep neural network~\cite{shaker2023swiftformer,transformer} and generative models~\cite{diffusionbeatgan,DDPM,DDIM,jiang2023talk}, many works have utilized neural networks to generate human motion~\cite{petrovich2021action,petrovich2022temos,petrovich2023tmr,zhang2021we,zhang2022wanderings,MDM,motiondiffuse,karunratanakul2024optimizing}. HumanML3D~\cite{guo2022generating} first created a high-quality text-motion paired dataset based on AMASS~\cite{mahmood2019amass} and designed an effective text-driven motion generation network based on VAE~\cite{autovae} and Transformer~\cite{transformer}. 
Following this approach, works such as TM2T~\cite{chuan2022tm2t}, T2M-GPT~\cite{t2m}, HumanTomato~\cite{lu2023humantomato} and MotionGPT~\cite{jiang2023motiongpt} encode motions into a series of discrete motion tokens using VQ-VAE~\cite{VQVAE}, and then employ a GPT~\cite{gpt1,gpt2,gpt3} based sequence model to learn the mapping between text tokens and motion tokens, achieving text-driven motion generation.
These framework encodes motions into latent space and then uses a sequence model to learn the dependencies between conditions and motions. 
The compressed latent space effectively reduces the complexity of sequence modeling, making it easier to generate logically coherent long-sequence motions. However, encoding motions into latent space reduces the diversity of the motions, and the pre-trained motion encoding and decoding networks limit further optimization of fine-grained motion quality.

\textbf{Diffusion models} ~\cite{DDPM} have achieved great success in tasks such as image~\cite{croitoru2023diffusion,yue2024resshift,yue2024efficient,yue2024resshift,liu2024residual}, speech~\cite{He_2024_CVPR}, video~\cite{shao2024human4dit}, and 3D generation~\cite{cai2021playing,li2024transformer,wang2024perf,xie2024gaussiancity,xie2024citydreamer,li2024animatable,xu2019unstructuredfusion}. 
Consequently, some works have also leveraged diffusion models for motion generation. 
MotionDiffuse \cite{motiondiffuse} first utilizes diffusion for the Text2Motion task, producing high-quality motion outputs; MDM~\cite{MDM} and MDM-Prior~\cite{MDMP} further enhance text-driven motion generation performance and enable diffusion-based motion editing, motion in-between, etc.
GestureDiffuCLIP \cite{gesturediffuclip} generates synchronized motions with speech and incorporates style control using text and video inputs.
% SAGA \cite{SAGA} and Grasping \cite{Grasping} specialize in natural grasping motion generation; \cite{DIMOS, zhang2022wanderings, huang2023diffusion} are capable of generating human motions that interact with 3D environments while avoiding collisions. CALM \cite{CALM} and ASE \cite{ASE} integrate reinforcement learning and physical simulation to improve the physical realism of generated movements. 
Additionally, the Large Motion Module~\cite{zhang2024large} trained a diffusion model that supports multimodal-driven motion generation, significantly improving motion generation quality and enabling its application to various tasks.
These diffusion-based motion generation methods have the advantages of strong controllability and high motion quality, but they struggle to model the logical relationships in long-sequence motions.

\textbf{Physical character animation} approaches~\cite{hassan2023synthesizing, deepmimic,xu2023adaptnet,wang2024pacer+,xiao2023unified,pan2024synthesizing} leverage physics engines to simulate realistic human motions by modeling the dynamics of the human body and its interactions with the environment. They employ reinforcement learning algorithms to learn motion skills. These methods ensure physical plausibility and can handle interactions with external forces or constraints.  However, physics simulation environments also significantly increase computational overhead, which limits the complexity of network models and the scale of datasets used.
Moreover, generating long dance sequences presents additional challenges. Long sequences involve complex and high-difficulty dance moves that necessitate training on extensive long-sequence dance datasets, further increasing computational demands. This makes it difficult to train long dance sequences using these methods. 
On the other hand, dance motions require intricate force techniques, and existing physical simulation environments typically model limbs as simple capsule shapes, which are insufficient for simulating complex dance dynamics. Although RFC~\cite{yuan2020residual} proposed the external residual forces to mitigate this issue, such approaches further increase the training difficulty and reduce the realism of the generated dance.

Compared to locomotion and fundamental actions such as running, long jumping, and squatting that are commonly studied in human motion generation, dance movements exhibit a higher degree of complexity.
Long sequence dances encompass not only extremely challenging poses but also meticulously choreographed sequences to enhance memorability and coordination, thereby more effectively conveying the dancer's emotions. Consequently, our method leverages state-of-the-art motion generation networks in combination with the choreographic processes of dancers. Furthermore, to address common issues in complex dance motions, such as sudden motion changes, foot-ground contact, and self-penetration, we develop dedicated modules that significantly enhance the overall quality and aesthetic appeal of the dance.

\subsection{Music Driven Dance Generation}
Research on generating dance sequences that are tightly synchronized with music input spans a variety of methods, including motion-graph approaches~\cite{choreographers}, sequence models~\cite{aist++, kim2022brand, bailando}, VQ-VAE models~\cite{bailando, gtn}, GAN-based methods~\cite{kim2022brand}, and diffusion-based techniques~\cite{edge, finedance}.

The motion-graph based dance generation methods~\cite{ofli2011learn2dance, choreonet, choreomaster} typically utilize a cross-modal retrieval network to calculate matching scores between music and dance segments.
Based on these matching scores and predefined choreography rules, a motion graph is constructed to synthesize long dance sequences. However, the predefined choreography rules are not adaptable to various dance genres, restricting these approaches to single-genre dances. Additionally, they are unable to achieve fine-grained beat alignment and lack diversity and creativity in the generated dances.

Deep learning approaches have become more widely adopted for generating high-quality, visually appealing dance movements. The most straightforward approach~\cite{huang2020dance, sun2022you} to generating dance using neural networks is to employ a sequence model such as LSTM~\cite{hochreiter1997long} or Transformer~\cite{transformer}. For example, FACT~\cite{aist++} employs a Transformer network to generate new dance frames conditioned on both music and initial motion frames. However, issues like error accumulation and motion freezing~\cite{music2dance} still arise.

Some methods use VQ-VAE~\cite{VQVAE} to convert dance motions into a sequence of dance tokens, then employ a sequence model~\cite{transformer} to learn dependencies between music and dance tokens. 
These methods~\cite{gong2023tm2d,zhuang2023gtn} can capture complex and vivid global choreography patterns. However, the two-stage design hinders the optimization of fine-grained dance quality, such as precise beat alignment and motion realism. Bailando~\cite{bailando} addresses the beat alignment issue by introducing reinforcement learning and a beat-align reward function, enabling the generated dances to balance global choreography patterns with fine-grained beat alignment.
Nevertheless, there is still considerable room for improvement in the fine-grained dance quality of Bailando~\cite{bailando}.

Generative Adversarial Networks (GANs)~\cite{goodfellow2014generative,radford2015unsupervised,bengio2013representation}, which involve a generator and discriminator in an adversarial structure, have been used to create realistic dance animations.
For instance, MNET~\cite{kim2022brand} employs a transformer-based generator along with a multi-genre discriminator to control dance genres. However, GAN-based techniques often struggle with challenges like mode collapse and training instability.

% More recently, diffusion models like FineDance~\cite{finedance} and EDGE~\cite{edge} have been employed to produce diverse, high-quality dance clips. However, due to their focus on generating detailed clips, these models are less effective at rapidly producing longer dance sequences with coherent choreography patterns.

More recently, diffusion models like FineDance~\cite{finedance} and EDGE~\cite{edge} have been employed to produce diverse, high-quality dance clips. Although FineDance incorporates Motion Graph to generate long-sequence dances, and EDGE maintains temporal consistency during the generation of long sequences, neither establishes a robust network for learning choreography patterns. Consequently, both methods exhibit some discontinuities at the transitions between dance segments.

In summary, while significant progress has been made in dance generation, producing high-quality, long dance sequences that maintain robust global choreography structures remains challenging.

%% file: sec/3_Methodology.tex
% \ifCLASSOPTIONcompsoc
% \IEEEraisesectionheading{\section{Method}\label{sec:method}}
% \else
\section{Method}
\label{sec:method}

\subsection{Preliminaries}
\noindent\textbf{Music and Dance Representation.} For music representation, we follow the approach in~\cite{aist++} and use Librosa~\cite{librosa} to extract the whole music feature $\bm{m}_w \in \mathbb{R}^{L \times 35}$, where $L$ represents the number of frames and 35 denotes the music feature channels, including 1-dim envelope, 20-dim MFCC, 12-dim chroma, 1-dim one-hot peaks, and 1-dim one-hot beats. 
For dance representation, we follow EDGE~\cite{edge} and define the whole dance sequence as $\bm{d}_w \in \mathbb{R}^{L \times 139}$, where the motion format adheres to SMPL~\cite{SMPL} and includes: (1) a 4-dimensional foot-ground contact binary label for left toe, left heel, right toe, and right heel, where 1 indicates contact with the ground and 0 indicates no contact; (2) 3-dim root translation; (3) 132-dim rotation data in a 6-dim rotation representation~\cite{6drot}, with the first 6 dimensions as global rotation and the remaining 126 for relative rotations of 21 sub-joints along the kinematic chain.

\noindent\textbf{The Diffusion Model.}
% We follow DDPM~\cite{DDPM} and EDGE~\cite{edge} to build our dance generation model. 
The diffusion model consists of a diffusing process and a denoising process. 
The diffusion process perturbs the ground truth dance data  $\bm{d}_0$ into $\bm{d}_t$ over $t$ steps, we follow \cite{DDPM} to simplify this multi-step diffusion process into one step, which can be formulated as:
\begin{equation}
  \begin{split}
            \begin{array}{lr}
             q\left(\bm{d}_{t}| \bm{d}_{0}\right) =\mathcal{N}(\sqrt{\bar{\alpha}_{t}}\bm{d}_{0},{(1-\bar{\alpha}_{t})} \bm{I}),
            \end{array}
  \end{split}
\end{equation}
where $\bar{\alpha}_{t}$ is within the range of $(0,1)$ and follows a monotonically decreasing schedule. 
$\bar{\alpha}_{t}$ converges to $0$ as $t$ goes to infinity, making $\bm{d}_{t}$ converging to a sample from the standard normal distribution.
The denoising process employs a Transformer base-network  $f_{\bm{\theta}}$ to gradually recover the motion, generating ${\hat{\bm{d}}}_0$ conditioned on given music $\bm{m}$. 
Instead of predicting the noise \cite{motiondiffuse}, we directly predict the ${\hat{\bm{d}}}_0$ like \cite{edge}. Therefore, the training process can be formulated as:
% \yan{to double check this formula} 
% \lrhP{Yes, I follow EDGE to write this formula.}

\begin{equation}
    \mathcal{L}_{\text{recon}}=\mathrm{E}_{\bm{d}_0, t}[\left\|{\bm{d}_0}-{f_{\bm{\theta}}}\left(\bm{d}_t, t, \bm{m}\right)\right\|_2^2].
\end{equation}

\subsection{Overview of Lodge++}
Given a long music feature $\bm{m}_w \in \mathbb{R}^{L\times 35}, L=ln$, and dance genre $g$, Our goal is to learn a neural network Lodge++, ${\bm{d}_w}=\textit{Lodge++}(\bm{m}_w,g), {\bm{d}_w}\in \mathbb{R}^{L\times 139}$.
We first use the Global Choreography Network to autoregressively generate a coarse dance sequence with $L$ frames quickly, and then extract dance primitives. Next, we split $\bm{m}_w$ into segments of length $n$ without overlaps, i.e. $\left\{{{\bm{m}}^i \in \mathbb{R}^{n\times35}}\right\}_{i=1}^{l}$. PDDM can parallelly generate $\left\{{{\bm{\hat{d}}}^i \in \mathbb{R}^{n\times139}}\right\}_{i=1}^{l}$ based on the given music clips and corresponding dance primitives, and finally get ${\bm{d}_w}=\textit{concatenate}( \lceil{{\bm{\hat{d}}}^i }\rceil,dim=0)$.

% Given a extremely long music feature $\bm{m} \in \mathbb{R}^{L\times 35}, L=kN={k^\prime}n$, we first split $\bm{m}$ into segments of length $N$ without overlaps, i.e. $\left\{{{\bm{m}}_w^i \in \mathbb{R}^{N\times35}}\right\}_{i=1}^{k}$.
% %\yan{with $L$ being normally larger than a number?}\lrh{Yes!L is larger than N} 
% %Then, we split each $\bm{m}_w$ into $\left\{{\bm{m}_l^j\in\mathbb{R}^{n\times35}}\right\}_{j=1}^{\lceil N/n \rceil}$.
% Our goal is to learn a neural network Lodge++, ${\bm{d}}_w^i=Lodge\text{++}({d}_w^i), {\bm{d}}_w\in \mathbb{R}^{N\times 139}$, ${\bm{d}}=concatenate( \lceil{{\bm{d}}_w^i }\rceil,dim=0)$,  which means Lodge++ can parallelly generate extremely long dance sequences ${\bm{d}} \in \mathbb{R}^{kN\times 139}$ with a single inference.

\begin{figure*}[t]
\begin{center}
\includegraphics[width=0.95\textwidth]{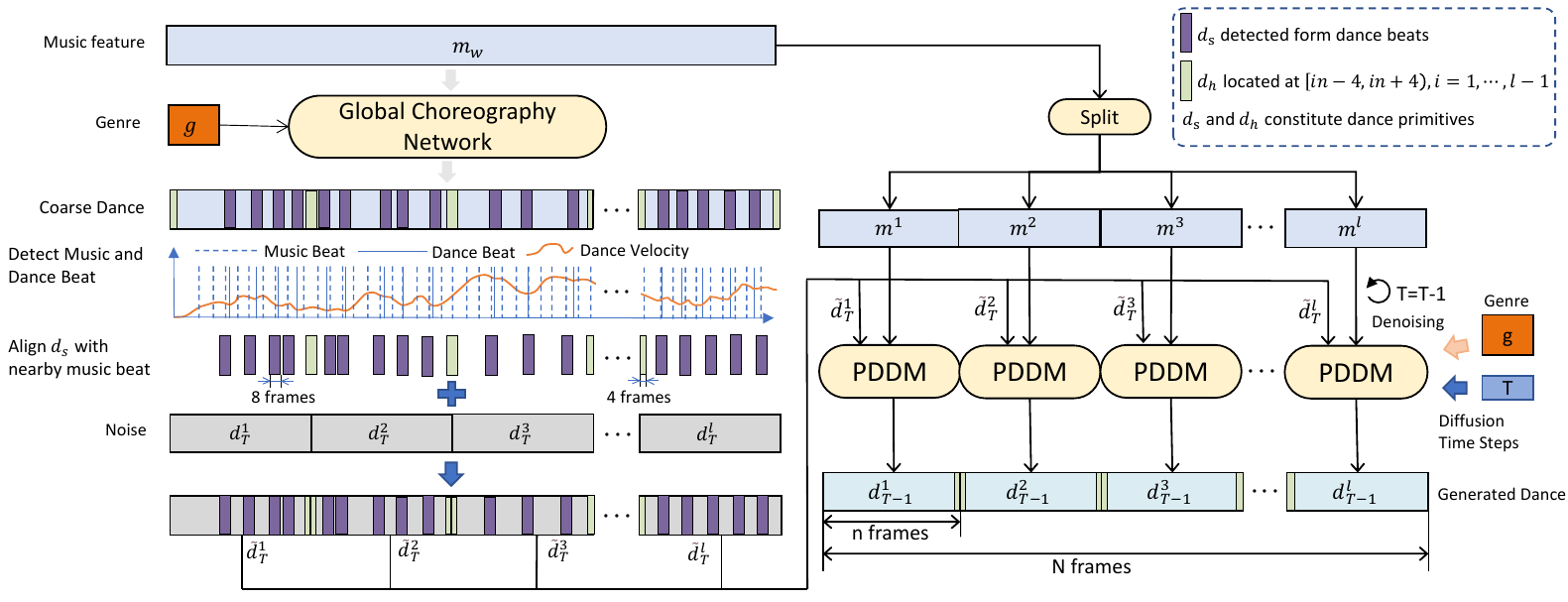}
% \vspace{-3mm}
\caption{The architecture of Lodge++. First, a Global Choreography Network is used to obtain coarse-grained dance motions. Then, expressive key motions near the dance beats of these coarse-grained motions are detected and aligned with their corresponding music beats, forming $\bm{d}_s$. These key motions serve to transfer the choreography patterns learned by the Global Choreography Network, further enhancing the expressiveness and beat alignment of the dances generated by PDDM.
The 8 frames of motion near $\left\{in\right\}_{i=1}^l$ are extracted as $\bm{d}_h$, which are used to constrain the start and end 4 frames of PDDM-generated local dance, supporting parallel generation in PDDM.
Both the $\bm{d}_s$ and $\bm{d}_h$ combine the dance primitives.
Next, noise is merged with dance primitives to obtain $\Tilde{\bm{d}}^i_T$. The $\Tilde{\bm{d}}^i_T$ and the split music features $\bm{m}^i$ are then input into the Primitive-base Dance Diffusion Model in parallel. After $T$ denoising steps, the final generated dance is obtained.}	
% \vspace{-5mm}
\label{fig:overallframework}
\end{center}
\end{figure*}

We find that employing a two-stage network that separately focuses on global choreography patterns and local dance quality is highly effective~\cite{li2024lodge}. However, the earlier version of this work, Lodge, modeled global choreography patterns by learning Characteristic Dance Primitives and applying basic choreographic rules. Due to the temporal sparsity of these Characteristic Dance Primitives and the oversimplified choreography rules, Lodge was unable to fully capture the richness and complexity of global choreography patterns.

Therefore, in Lodge++, we learn global choreography patterns in a compact  implicit space. Additionally, we need to transfer the Dance Primitives learned by the global choreography network to the local dance generation network. 
The diffusion guidance method used in Lodge introduces a gap between training and inference, sometimes leading to unnatural movements, such as abrupt transitions.
To address this, we design a new Primitive-based Dance Diffusion network in Lodge++.

The pipeline of Lodge++ is shown in the Figure~\ref{fig:overallframework}. Given music features $\bm{m}_w$ and the desired dance genre $g$, the Global Choreography Network first generates coarse-grained dance. We then detect dance beats based on the minima of dance movement speed and extract 8 frames of dance near each dance beat as 
$\bm{d}_s$, and frames $[in-4, in+4)$, where $i=1,2,3,\cdots,l-1$, as $\bm{d}_h$. Together, 
$\bm{d}_s$ and $\bm{d}_h$
constitute the dance primitives, which are used to convey overall choreography patterns.

Finally, we segment $\bm{m}$ into 
$\left\{{\bm{m}^i}\right\}_{i=1}^{l}$ over time and merge the pure noise $\bm{d}^i_T$ with the dance primitives to get ${{\hat{\bm{d}}}^i_T}$. After $T$ denoising steps, we generate 
$\left\{{\hat{\bm{d}}^i_0}\right\}_{i=1}^l$. Since 
$\bm{d}_h$ is used to constrain the initial and final frames of $\bm{d}^i_0$, PDDM supports parallel generation, greatly improving generation efficiency.

\subsection{Global Choreography Network}

Long-sequence dances involve complex choreography patterns, and the choreography rules vary across different dance genres.
Previous work~\cite{aist++,finedance,sun2022you,luo2024popdg,li2022danceformer,wang2024dancecamera3d,wang2022groupdancer} overlooked the learning of global choreography patterns, resulting in long sequences of dance that are chaotic and disordered.

To alleviate the substantial computational load brought by long sequence data, we first encode and quantize the coarse-level dance of main joint rotation movements into a Choreographic Memory Codebook using VQ-VAE. Each segment of the dance can be encoded into a series of tokens. 
The coarse-level dance is robust enough to express rich choreography patterns.
To learn the complex global choreography patterns between music and dance, we use a sequence model, specifically a Transformer, to learn the dependencies between the music and dance tokens.

\subsubsection{Dance VQ-VAE for Choreography Memory Encoding}

Dance is composed of many basic movement elements. By arranging and combining these basic elements with appropriate modifications, a rich variety of long-term dances can be choreographed. Inspired by this idea, Bailando and Bailando++ use VQ-VAE to encode dances into a Choreographic Memory Codebook, which can encode a short segment of dance into discrete Dance Tokens. To further enhance the capacity of the Choreographic Memory Codebook, they decouple upper-body and lower-body movements. 
Despite showing significant progress, existing VQ-VAE networks still struggle to precisely compress and reconstruct full-body dance movements due to the inherent diversity and complexity of dance movements.
This is especially true for challenging movements like fast turns and backflips, which often result in unrealistic reconstructed motions.

% We draw inspiration from Labanotation, which demonstrates that the primary body joints are sufficient to record global choreography patterns.
We lower the demands on VQ-VAE by encoding only the rotation angles of the main body joints: specifically, the rotations of the root, left shoulder, right shoulder, left arm, right arm, left leg, and right leg. Therefore, we extract the main body joints from each dance sequence $\bm{d}\in\mathbb{R}^{N\times 139}$ in the dataset to obtain $\bm{d}^\prime\in\mathbb{R}^{N\times 42}$, where $N$ is the time length, 139 is the feature dimension of origin dance, 42 is the 6-dim rotation information of the above 7 main joints.
Then, we use these coarse-grained dance $\bm{d^\prime}$ to train a VQ-VAE network.
The VQ-VAE network maintains a learnable choreography memory codebook. The VQ-Encoder $\mathcal{E}$ can map any sequence of coarse dance to a series of discrete tokens in the codebook, and the VQ-Decoder $\mathcal{D}$ can reconstruct high-quality coarse dance from these discrete tokens.

Specifically, the VQ-VAE encoder use 1D temporal convolution network to convert coarse-grained dance $\bm{d}^{\prime}\in\mathbb{R}^{N \times 42}$ into 
$\left\{{{\hat{\bm{z}}_1}},{\hat{\bm{z}}_2}, \cdots ,{\hat{\bm{z}}_{N^\prime}}\right\},$
% $\hat{Z}=\left\{{\hat{z}_i}\right\}_{i=1}^{T^\prime},\hat{z}_i\in\mathbb{R}^{C_z}$
% ${\hat{z}\in\mathbb{R}}^{T^\prime\times {C_z}}$
where $N^\prime=N/r$, $r$ is the temporal down-sampling rate, ${\hat{\bm{z}}_i}\in\mathbb{R}^{C_z}$, $C_z$ is the channel dimension. 
Then we quantize $\hat{z}$ into $z$ through searching the nearest embedding in a learnable choreography memory codebook $Z=\left\{{{\bm{z}_i}}\right\}_{i=1}^K,{\bm{z}_i}\in\mathbb{R}^{C_z}$, formulated as:
\begin{equation}
\bm{z}_i = \underset{\bm{z}_j\in Z}{\operatorname{argmin}} \left\| {\hat{\bm{z}}_i} - \bm{z}_j \right\|
\end{equation}

The VQ-VAE Decoder can decode the quantized features $z$ and reconstruct the dance movement. The optimization objective of the VQ-VAE model is as follows:
\begin{equation}
\begin{aligned}
\mathcal{L}_{VQVAE}&=\left\|\mathcal{D}(\mathcal{E}(\bm{d}^{\prime}))-\bm{d}^{\prime}\right\|\\
&+\left\|sg\left[\mathcal{E}(\bm{d}^{\prime})\right]-\bm{z}\right\|+\beta \left\|\mathcal{E}(\bm{d}^{\prime})-sg\left[\bm{z}\right]\right\|,
\end{aligned}
\end{equation}
where $\mathcal{E}$ represents the VQ-VAE Encoder, $\mathcal{D}$ represents the VQ-VAE Decoder, and $sg\left[\cdotp\right]$ denotes parameter freezing, $\beta$ is a hyper-parameter for loss balance. 
% The first term is used to train the encoder and decoder, the second tem

\begin{comment}
  舞蹈由很多基本的动作元素构成，通过对这些基本动作元素进行排列组合以及适当的修改，可以编排出丰富多样的长时舞蹈。受这一思路启发，Bailando和Bailando++采用VQ-VAE将舞蹈编码到Choreographic Memory Codebook，可以把一小段舞蹈编码成离散的Dance Token。为了进一步提升Choreographic Memory Codebook对舞蹈动作的表达能提，他们还将上身和下身动作解耦。Bailando和Bailando++提出的VQ-VAE展现出了优秀的舞蹈压缩和重建能力。然而，由于舞蹈动作本质上的丰富多样性和挑战性，现有的VQ-VAE网络难以精准地压缩并重构出全身的舞蹈动作，特别是一些难例动作如快速转体，后空翻等动作会出现不真实的重建动作。
  为了解决这个问题，兼顾我们的由粗到细的层次式网络设计，我们降低了对VQ-VAE要求，只用VQ-VAE编码主要的身体关节的旋转角，分别是（root，左肩膀、右肩膀、左臂、右臂、左腿、右腿）这7个主要身体部位的旋转角度。
  因此，我们从原始舞蹈数据d中提取上述主要身体部位的旋转数据，获得粗粒度舞蹈d^prime。
  早期的工作采用Motion Graph的方法~\cite{ChoreoMaster}，对数据库中的舞蹈动作按照一些基本编舞规则进行裁剪和重组。
  然而，这类方法一方面
\end{comment}

\begin{comment}
  一个完整的舞蹈包含着丰富的舞蹈结构，通过恰当的动作重复、转折或递进，使整个作品更加生动和有层次感。之前的工作忽略了舞蹈的全局编舞规律的学习，导致长序列的舞蹈动作混乱和无序。
  
  为了减轻长序列数据带来巨大的计算量，我们将舞蹈动作先用VQ-VAE编码并量化到连续的Choreographic Memory Codebook，每段舞蹈可以被编码成一连串的Tokens；为了学习复杂的音乐和舞蹈之间的全局编舞规律，我们利用一个序列模型Transformer学习音乐和舞蹈Tokens的依赖。

\end{comment}

\begin{figure*}[t]
\begin{center}
\includegraphics[width=1.0\textwidth]{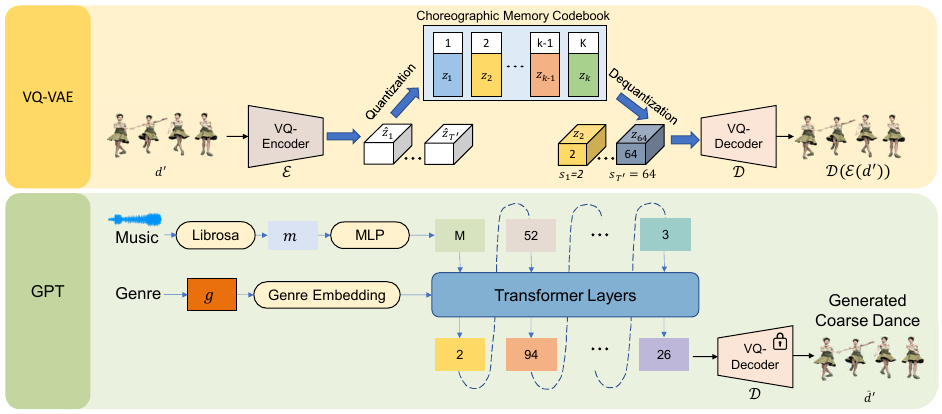}
% \vspace{-3mm}
\caption{Structure of Global Choreography Network. The Global Choreography Network consists of two parts: VQ-VAE and a sequence model. First, VQ-VAE encodes the dance movements into a choreography memory codebook. Then, the sequence model generates coarse-grained dance that adheres to choreography patterns based on the input music and dance style.}	
% \vspace{-5mm}
\label{fig:framework}
\end{center}
\end{figure*}

\begin{comment}
我们发现采用两阶段网络，分别关注于全局编舞规律和局部舞蹈质量是有效的~\cite{Lodge}。
然而，这个工作的早期版本Lodge先通过学习稀疏的Characteristic Dance Primitives来建模全局编舞规律，但是由于Dance Primitives在时域上过于稀疏，导致其无法充分表达丰富的全局编舞规律。
因此在Lodge++中，我们在一个压缩但连贯的隐式空间学习全局编舞规律。另外，我们需要将全局编舞网络学到的Dance Primitives传递到局部舞蹈生成网络，之前的Lodge采用diffusion guidance的方法，会引入训练和推理的gap，造成局部舞蹈生成网络有时无法根据具有表现力的舞蹈动作。因此，在Lodge++中，我们设计新的残差diffusion网络。
Lodge++的pipeline如图所示，给定音乐特征mg和想要生成的舞蹈风格g，首先Global Choreographt Netword生成粗粒度的舞蹈。我们随后根据舞蹈运动速度的极小值点检测舞蹈节拍，并且将dance beat附近的8帧舞蹈提取为作为ds，将第[ln-4，ln+4)，其中l=1，2，3，4，4n=N，提取为dh，ds和dh共同构成Characteristic Dance Primitives，用于表达整体编舞规律。
最后，我们将mg按时间切分为ml^i,将原始的Diffusion网络的噪声dl替换为叠加了Characteristic Dance Primitives的dl’，经过T个去噪步后，生成dl0。由于dh用于约束RDM的首尾帧动作，因此RDM支持并行生成，极大地提高了生成效率。

我们发现长序列舞蹈生成的主要计算开销来自于对全局编舞规律的建模需求。因此需要对全局音乐和舞蹈特征进行压缩来学习，这个工作的早期版本Lodge采用稀疏的Characteristic Dance Primitives来表达全局编舞规律，但是由于Dance Primitives在时域上过于稀疏，导致其无法充分表达丰富的全局编舞规律。因此我们在一个压缩的隐式空间学习全局编舞规律，然后利用RDDM的设计将其传递到Local Diffusion网络。Local Diffusion关注于生成高质量且物理真实的短时舞蹈片段，且通过并行生成的架构可以大幅提升长序列生成的效率。
\end{comment}

\subsubsection{Motion GPT for Global Choreography Learning}

Thanks to the Dance VQ-VAE model, coarse-grained dance sequences $[{\bm{d}^{\prime}_1}, \bm{d}^{\prime}_2, ..., \bm{d}^{\prime}_{T}]$ can be transformed into $[\bm{z}_1, \bm{z}_2, ..., \bm{z}_{T^{\prime}}]$. Due to the VQ-VAE being pre-trained, we can further rewrite $[\bm{z}_1, \bm{z}_2, ..., \bm{z}_{T^{\prime}}, End]$ into a series of tokens $S=[s_1, s_2, ..., s_{T^{\prime}}, End]$, $s_i$ is the index of $z_i$ in the choreography memory codebook, $End$ is a special token means that the sequence is ended.

% a series of tokens from the choreography memory codebook: $[z_1, z_2, ..., z_{T^{\prime}}, End]$ 
% , where $
  % represents an index in the codebook. 
% By modeling the mapping between music and the dance tokens using a GPT model, we can efficiently learn the knowledge required for global choreography. 
Consequently, global choreography can be formed as a sequence prediction problem. Given the music feature $\bm{m}$ and dance genre $g$, the GPT autoregressively predicts the next dance token, formulated as:

\begin{equation}
\begin{split}
p_{\bm{\theta}}\left(s_t\middle|\bm{m},g\right) &=p_{\bm{\theta}}\left(s_t\middle|\bm{m},g,s_{<t}\right),
\\
p_{\bm{\theta}}\left(S\middle|\bm{m},g\right)&=\prod_{i}^{\left|S\right|}{p_{\bm{\theta}}\left(s_i\middle|\bm{m},g,s_{<i}\right)}
\end{split}
\end{equation}

The optimization goal of dance GPT Network is:

\begin{equation}
\mathcal{L}_{GPT}=E_{S\sim P_{\bm{\theta}}(S)}[-log{\ p}_{\bm{\theta}}\left(S\middle| \bm{m},g\right)]
\end{equation}

Based on GPT and the choreography memory codebook, our proposed Global Choreography Network learns complex and vivid  global choreography patterns, making it adaptable to a wide range of dance genres.
% Unlike Lodge's approach, which heavily relies on manually predefined choreography rules, this method, based on GPT and the choreography memory codebook, can learn more robust and complex global choreography patterns, making it applicable to a broader range of music and dance styles.
\begin{comment}
  得益于Dance VQ-VAE模型，可以将粗粒度的舞蹈序列[x1, x2, . . . , xT ]，转化为一系列编舞记忆codebook里面的Tokens:[z1, z2, . . . , zT',End],zi为codebook中的索引。通过用GPT模型建模音乐与舞蹈tokens之间的映射，可以高效地学习根据音乐进行全局编舞的知识。由此，我们可以将全局编舞转化为序列预测问题。根据给定音乐，和舞蹈流派，GCN自回归地预测下一个token，公式为
  
  不同于Lodge过多地采用人工设定的编舞规则，这种基于GPT和编舞记忆codebook的方法可以学到更鲁棒和更复杂的全库编舞规律，适用于更广泛的音乐和舞蹈类型。
\end{comment}
However, the global choreography network faces challenges with beat alignment due to the token compression in the latent space of VQ-VAE. 
% uses a VQ-VAE+GPT architecture.
To address this issue, we first use the VQ-Decoder to reconstruct coarse-grained dance movements from the predicted sequence of dance tokens and then detect the motion beats within coarse dance. Next, we extract the 8 frames surrounding each dance beat and align them with the nearest music beats, resulting in $d_s$. Additionally, we extract the motion frames  $[in-4,in+4),i=1,\cdots,l-1$ as $\bm{d}_h$.
Both the $\bm{d}_s$ and $\bm{d}_h$ consist of the dance primitives, where $\bm{d}_s$ can convey the complex choreography patterns learned by the Global Choreography Network, guiding the local PDDM to generate more expressive dance movements, while $\bm{d}_h$ can be used to support parallel generation.
% This approach not only captures rich choreography patterns but also improves the alignment of the generated dance with the given musical beats.

\subsection{Primitives-based Dance Diffusion Model}
\label{sec:PDDM}

\begin{figure}[t]
\begin{center}
\includegraphics[width=0.45\textwidth]{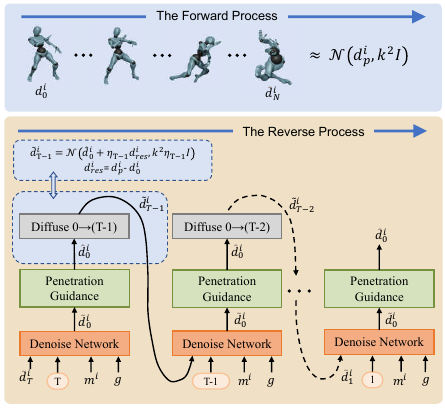}
% \vspace{-3mm}
\caption{The forward and reverse process of the Primitive-based Dance Diffusion Model. $\mathcal{T}$ is the diffusion time steps.}	
% \vspace{-6mm}
\label{fig:framework}
\end{center}
\end{figure}

To facilitate the transfer of dance primitives from the Global Choreography Network to the local dance generation network and to minimize the gap between training and inference, we propose the Primitives-based Dance Diffusion Model (PDDM).
The PDDM can generate high-quality local dance clips  $\left\{{{\hat{\bm{d}}_0^i} }\right\}_{i=1}^{l}$ based on dance primitives $\left\{{{{\bm{d}^i_p}} }\right\}_{i=1}^{l}$, the given music $\left\{{{\bm{m}}^i }\right\}_{i=1}^{l}$, and the desired dance genre $g$.
${\hat{\bm{d}}_0^i}=\text{PDDM}(\bm{m}^i,{\bm{d}^i_p},g)$.
Based on the $\bm{d}_h$ in dance primitives, the PDDM supports parallel generation to enhance the efficiency. 
For the sake of simplicity, we omit the superscript $i$ in section~\ref{sec:process} and section~\ref{sec:losses}.

\subsubsection{The forward and reverse process of PDDM}
\label{sec:process}

\textbf{The forward process} 
% $d_0$ and $d_p$ to $d_T$ 
is to progressively shift the residual ${\bm{d}_{res}}={\bm{d}_p}-\bm{d}_0$ and inject noise.
The forward process of the PDDM can be formulated as:

% \vspace{-3mm}
\begin{align}
    q({\bm{d}_{t}}|{\bm{d}_{t-1}},{\bm{d}_0},{\bm{d}_p})&=\mathcal{N}({\bm{d}_t};{\bm{d}_{t-1}}+\alpha_t {\bm{d}_{res}},k^2 \alpha_t \bm{I}),\\
    q({\bm{d}_{1:T}}|{\bm{d}_0},{\bm{d}_{p}})&=\prod_{t=1}^{T}{q(\bm{d}_t|\bm{d}_{t-1},\bm{d}_{res})},   \\
    q(\bm{d}_{t}|\bm{d}_0,\bm{d}_{p})&=\mathcal{N}(\bm{d}_t;\bm{d}_{0}+\eta_t \bm{d}_{res},k^2 \eta_t \bm{I}),
\end{align}
where $\alpha_t$ is a hyper-parameter controlling the shifting trajectory. The smaller $\alpha_t$ is, the smoother the shifting becomes.
$k$ is a hyper-parameter controlling the noise variance. $\alpha_t=\eta_t-\eta_{t-1}$ for $t>1$,$\eta_1=\alpha_1\rightarrow 0,\eta_T=\sum_{i=1}^T \alpha_t\rightarrow 1$. Therefore, $q(\bm{d}_{T}|\bm{d}_0,\bm{d}_{p})$ converges to $\mathcal{N}(\bm{d}_T;\bm{d}_p,k^2 \bm{I})$, which is  approximate distribution for the dance primitives $\bm{d_p}$.

% The forward process means the residual $d_{res}$ and noise $\epsilon$ are gradually added to $d_0$. 

\textbf{The reverse denoising process} is to recover $\bm{d}_0$ from $\bm{d}_p$ and $\bm{c}$, where $\bm{c}$ is the condition derived by concatenating music $\bm{m}$ and genre $g$.
In Lodge++, we utilize a denoising network $p_{\bm{\theta}}\left(\bm{d}_{t-1}|\bm{d}_t,\bm{d}_p,\bm{c}\right)$ to estimate the posterior distribution $p(\bm{d}_0|\bm{d}_T,\bm{d}_{p},\bm{c})$:
% \end{comment}
\begin{equation}
    p(\bm{d}_0|\bm{d}_T,\bm{d}_p,\bm{c})=\int d_T \prod_{t=1}^{T} p_{\bm{\theta}}\left(\bm{d}_{t-1}|\bm{d}_t,\bm{d}_p,\bm{c}\right) \mathrm{d}_{\bm{d}_{1:T}},
\end{equation}
where $\bm{d}_{T}\approx\mathcal{N}(\bm{d}_T;\bm{d}_p,k^2 \bm{I})$. Following most of previous diffusion relative papers ~\cite{diff1,diff2,diff3}, we adopt the following assumption:
\begin{equation}
p_{\bm{\theta}}(\bm{d}_{t-1}\mid \bm{d}_t,\bm{d}_p,\bm{c})=\mathcal{N}\left( \bm{d}_{t-1};\mu_{\bm{\theta}}(\bm{d}_t,\bm{d}_p,\bm{c},t),\Sigma_{\bm{\theta}}(\bm{d}_t,\bm{d}_p,\bm{c},t)\right).
\label{eq:targetpddmdenoise}
\end{equation}

\subsubsection{The Training Objective of PDDM}
\label{sec:losses}
To optimize the neural network, we minimize the negative evidence lower bound~\cite{diff1,diff2}, specifically:
\begin{equation}
\label{eq:originobj}
\min_{{\bm{\theta}}}{\sum_{t}{D_{KL}(\left[q(\bm{d}_{t-1}|\bm{d}_t,\bm{d}_0,\bm{d}_p,\bm{c})\|p_{\theta}(\bm{d}_{t-1}|\bm{d}_t,\bm{d}_p,\bm{c})\right])}},
\end{equation}
where $D_{KL}\left[\cdotp\|\cdotp\right]$ represents the Kullback-Leibler (KL) divergence~\cite{kullback1951information}.
The targeted distribution $q(\bm{d}_{t-1}|\bm{d}_t,\bm{d}_0,\bm{d}_p)$ in Eq.(\ref{eq:targetpddmdenoise}) can be made tractable and represented as:
\begin{equation}
    q(\bm{d}_{t-1}|\bm{d}_t,\bm{d}_0,\bm{d}_p,\bm{c})=\mathcal{N}(\bm{d}_{t-1}|\frac{\eta_{t-1}}{\eta_t}\bm{d}_{t}+\frac{\alpha_t}{\eta_t}d_0,k^2\frac{\eta_{t-1}}{\eta_t}\alpha_t \bm{I}),
\end{equation}
where the variance parameter is independent of $\bm{d_t}$ and $\bm{d_p}$. We set the mean term as $\mu_{\bm{\theta}}(\bm{d}_t,\bm{d}_p,t)$, specifically:
\begin{equation}
\label{eq:mean}
    \mu_{\bm{\theta}}(\bm{d}_t,\bm{d}_p,\bm{c},t)=\frac{\eta_{t-1}}{\eta_t}\bm{d}_t+\frac{\alpha_t}{\eta_t}f_{\bm{\theta}}(\bm{d}_t,\bm{d}_p,\bm{c},t),  
\end{equation}
where $f_{\bm{\theta}}$ is a deep neural network to predict $\bm{d_0}$. Based on Eq.(\ref{eq:mean}), the objective function in Eq.(\ref{eq:originobj}) can be simplified to:
\begin{equation}
% \min_{\theta}{\sum_{t} \|f_{{\bm{\theta}}}(d_t,d_p,c,t)-d_0\|_2^2}.
\mathcal{L}_{recon}=\|f_{{\bm{\theta}}}(\bm{d}_t,\bm{d}_p,\bm{c},t)-\bm{d}_0\|_2^2.
\end{equation}

\subsubsection{The Auxiliary Losses}
In addition to reconstruction loss  $\mathcal{L}_{recon}$, we follow previous works ~\cite{li2024lodge, edge,lemo,xiang} by adding extra losses to enhance training stability and reduce motion jitter and foot sliding issues:
% \vspace{-1.5mm}
\begin{equation}
\begin{aligned}
{\bm{d}}_{\text{joint}}^{(j)} &= FK(\bm{d}^{(j)} ), %,    \\
% {\textbf{d}}_{\text{j-vel}}^{(i)} &= {\textbf{d}}_{\text{joint}}^{(i+1)} - {\textbf{d}}_{\text{joint}}^{(i)},   \\
% {\textbf{d}}_{\text{j-acc}}^{(i)} &= {\textbf{d}}_{\text{j-vel}}^{(i+1)} - {\textbf{d}}_{\text{j-vel}}^{(i)}
\end{aligned}
\label{eq:joints}
% \vspace{-1.5mm}
\end{equation}

% \vspace{-1.5mm}
\begin{equation}
\begin{aligned}
\mathcal{L}_{\text{joint}}=\frac{1}{n} \sum_{j=1}^{n} \left\| {\bm{d}}_{\text{joint}}^{(j)} - {\hat{\bm{d}}}_{\text{joint}}^{(j)} \right\|^2_2,
\end{aligned}
\label{eq:jointsloss}
% \vspace{-1.5mm}
\end{equation}

% \vspace{-1.5mm}
\begin{equation}
\begin{aligned}
\mathcal{L}_{\text{j-vel}}=\frac{1}{n-1} \sum_{i=1}^{n-1} \left\| {\bm{d}}_{\text{j-vel}}^{(j)} - {\hat{\bm{d}}}_{\text{j-vel}}^{(j)} \right\|^2_2,
\end{aligned}
\label{eq:vel}
% \vspace{-1.5mm}
\end{equation}

% \vspace{-1.5mm}
\begin{equation}
\begin{aligned}
\mathcal{L}_{\text{j-acc}}=\frac{1}{n-2} \sum_{j=1}^{n-2} \left\| {\bm{d}}_{\text{j-acc}}^{(j)} - {\hat{\bm{d}}}_{\text{j-acc}}^{(j)} \right\|^2_2,
\end{aligned}
\label{eq:acc}
% \vspace{-1.5mm}
\end{equation}
% \vspace{-1.5mm}
\begin{equation}
\begin{aligned}
\mathcal{L}_{\text{contact}}=\frac{1}{n-1} \sum_{j=1}^{n-1} \left\|  ({\hat{{\bm{f}}}}_{{hv}}^{(j)} + {\hat{{\bm{f}}}}_{{dv}}^{(j)})  \cdot {\hat{b}^{(j)}} \right\|^2_2,
\end{aligned}
\label{eq:contact}
% \vspace{-1.5mm}
\end{equation}
where $j$ means the $j-th$ frame, $n$ is the number of motion frames. ${\hat{b}}$ is the predicted foot contact label. $f_{hv}$ and $f_{dv}$ represent horizontal downward velocity and vertical velocity of feet, respectively.
The overall training objective is determined by combining the losses with weighted factors:
% \vspace{-1.5mm}
\begin{equation}
\begin{aligned}
\mathcal{L}_{\text{total}} =&\mathcal{L}_{\text{recon}}  
+ {\lambda}_{\text{joint}} \mathcal{L}_{\text{joint}} 
+ {\lambda}_{\text{j-vel}} \mathcal{L}_{\text{j-vel}} \\
&+{\lambda}_{\text{j-acc}} \mathcal{L}_{\text{j-acc}} 
+ {\lambda}_{\text{contact}} \mathcal{L}_{\text{j-contact}}
+ {\lambda}_{\text{genre}} \mathcal{L}_{\text{genre}}.
\end{aligned}
\label{eq:total}
% \vspace{-1.5mm}
\end{equation}
where $\lambda$ represents the weights for the corresponding loss terms.

\subsubsection{The Detailed Network Architecture of PDDM}
\begin{figure*}[t]
    \centering
    \includegraphics[width=\textwidth]{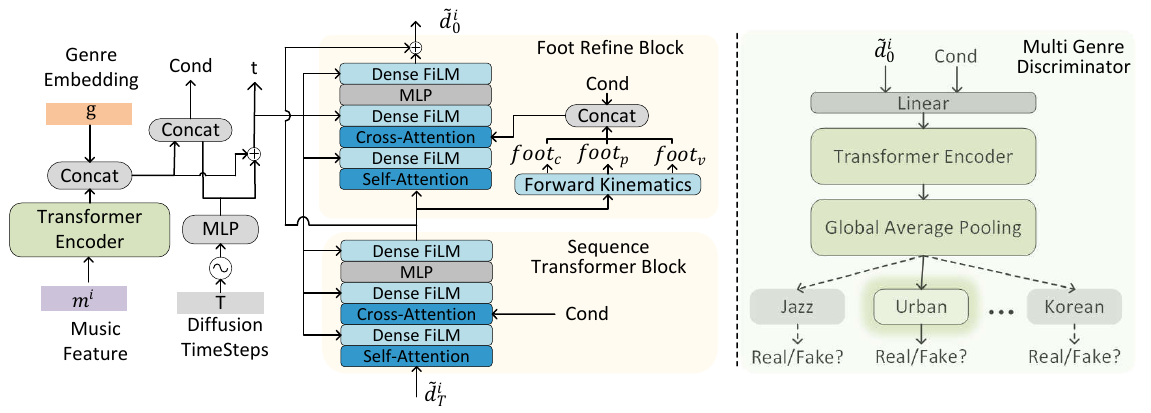}%\vspace{-4mm}
    \caption{The detailed network architecture of the Denoise Network in the Primitive-based Dance Diffusion Model.}%\vspace{-3mm}
    \label{fig:LocalDiffusion}
    % \vspace{-1.5em}
\end{figure*}

\noindent\textbf{Foot Refine Block.} In the SMPL format, motion representation is defined by relative rotations propagated along the kinematic chain, which means that minor change near root joints can cause large shifts at distal joints, such as the knees and feet. This issue is particularly challenging in dance movements, which often involve intricate foot actions, making it difficult to resolve foot-ground contact issues (e.g., foot sliding, foot floating) by simply applying foot-specific loss functions.

The primary problem arises from the domain gap between the optimization target and the data representation. Foot-ground contact is measured in a linear space based on joint positions, whereas SMPL-based motion exists within a nonlinear rotational space. To address this, we introduce a Foot Refine Block, which first calculates the positions of foot keypoints $\bm{foot}_p$ and foot velocity $\bm{foot}_v$ through forward kinematics.
% inspired by \cite{xiang}.  
Following \cite{xiang}, we then compute the foot-ground contact score $\bm{foot}_c$. A Cross Attention mechanism is subsequently applied to further refine foot movements. 

\noindent\textbf{Multi Genre Discriminator.} 
The PDDM can produce high-quality, diverse dance segments.
As shown in the Figure~\ref{fig:LocalDiffusion},
to ensure consistency with the overall dance genre, we also concatenate the genre embedding $g$ with the music features for feature encoding. We then use a Multi-Genre Discriminator (MGD) to control the dance genre following  MNET~\cite{kim2022brand}. The training process of MGD is formulated as:
% \vspace{-1.5mm}
\begin{equation}
\begin{aligned}
&\mathcal{L}_{\text {genre }}=  \mathbb{E}_{{\Tilde{\bm{d}}_0^i}}
\left[\log MGD\left({{\Tilde{\bm{d}}_0^i}}, g,{\bm{m}^i}\right)\right]+ \\
&\mathbb{E}_{{\Tilde{\bm{d}}_0^i} ,  t}\left[\log \left(1-     
MGD\left(LD\left({\Tilde{\bm{d}}_0^i}, g, \bm{m}^i\right), g, \bm{m}^i\right)\right)\right],
\end{aligned}
\label{eq:genre}
% \vspace{-1.5mm}
\end{equation}

\subsubsection{The Penetration Guidance}
We find that the hands in the generated dances often penetrate the body. To further enhance physical realism and address the penetration issue, we propose a Penetration  Guidance method based on a diffusion denoising process. 
By using the COAP~\cite{mihajlovic2022coap} model to compute the Signed Distance Field (SDF) between the hand vertices and other body vertices, we can obtain gradients to guide hand and arm movements to alleviate the penetration problem:
\begin{equation}\mathcal{G}_{pene}({\Tilde{\bm{d}}^i_t})=\frac{1}{|{\bm{p}}_h^n|}\sum_{q\in{\bm{p}}_h^n} \sigma\big(f_{{\bm{\theta}}}(q|\mathcal{S})\big)\mathbb{I}_{f_{{\bm{\theta}}}({\bm{p}}_h^n|\mathcal{S})>0},
\end{equation}
where $\bm{p}_h^i$ is the point set of hands calculated by ${\Tilde{\bm{d}}^i_t}$ using SMPL-X Skinning Function~\cite{SMPL-X}.
$f_{{\bm{\theta}}}(q|\mathcal{S})$ is the COAP model used to compute the signed distance between point $q$ and the body mesh $\mathcal{S}$, $\sigma(\cdot)$ is the sigmoid function.

\begin{equation}
{{\hat{\bm{d}}^i_t}}={\Tilde{\bm{d}}^i_t} + a_{con}\nabla\mathcal{L}_{con}({\Tilde{\bm{d}}^i_t}) + a_{pene}\nabla\mathcal{G}_{pene}({\Tilde{\bm{d}}^i_t}),
\end{equation}
where $a_{con}$ and $a_{pene}$ are scale factors and ${{\hat{\bm{d}}^i_t}}$ is the motion representation after guidance.

\subsubsection{Parallel Inference.}
The PDDM can generate ${\left\{{{\hat{\bm{d}}_0^i} }\right\}}_{i=1}^{l}$ in parallel. However, directly concatenating $\hat{\bm{d}}_0^i$ along the temporal axis to form a long dance sequence results in abrupt transitions at  every 
$(i\times n)-th$ frames, where 
$i=,\cdots,l-1$. To address this issue, we constrain ${\bm{d}^i}$ using $\bm{d}_h$ from the dance primitives.
% Given the inputs $\bm{m}^i$, $g$, and the corresponding $\bm{d}_h$ and $\bm{d}_s$, the local diffusion process generates $\bm{d}^i$. By concatenating $\left\{{\bm{d}^i}\right\}_{i=1}^l,$ along the time axis, we obtain the $\bm{d}_w$ as output. 
we divide each $\bm{d}_h$ into the first four frames and the last four frames. The first four frames serve as the tail four frames of the previous $\bm{d}^i$, and the last four frames of $\bm{d}_s$ serve as the leading four frames for the next $\bm{d}^{i+1}$. 

However, directly using diffusion inpainting~\cite{repaint} to control the first and last frames of each segment still results in incoherent motions. 
Therefore, we use a joint acceleration loss 
$\mathcal{L}_{\text{j-acc}}$ and modified the $\mathcal{L}_{\text{recon}}$ loss of the PDDM.
At each diffusion time step,  we mixture ${\Tilde{\bm{d}}^i_t}$ of and
the ground truth $\bm{{d}_0^i}$ by
$\bar{\bm{d}}^i_t[:4] = {{\bm{d}}_0}[:4]$, 
$\bar{\bm{d}}^i_t [-4:] = {\bm{d}_0}[-4:]$,
$\bar{\bm{d}}^i_t [4:-4] = {\Tilde{\bm{{d}}}_t}[4:-4]$.
The $\mathcal{L}_{\text{recon}}$ loss is formulated as:
% \vspace{-2mm}
\begin{equation}
\begin{aligned}
    \mathcal{L}_{\text{recon}} = \mathrm{E}_{{\bm{d}^i_0}, t}[\left\|{{\bm{d}^i_0}} - {f_{\bm{\theta}}}\left(\bar{\bm{d}}^i_t, g,t, \bm{m}^i\right)\right\|_2^2].
% \vspace{-3mm}
\end{aligned}
\end{equation}

%% file: sec/4_Experiments.tex
\section{Experiments}
\label{sec:experiments}

\subsection{Experimental Setup}
\textbf{Dataset:} We use the FineDance~\cite{finedance} dataset to train and validate our algorithm. Compared to the earlier publicly available music-dance dataset AIST++~\cite{aist++}, FineDance is performed by professional dancers and recorded with advanced optical motion capture equipment, offering better artistic quality and motion fidelity, and includes a richer variety of 22 dance genres. Notably, the \textbf{average duration} of each dance in FineDance is 152.3 seconds, significantly longer than AIST++'s 13.3 seconds. Therefore, FineDance is more suitable for evaluating long-sequence dance generation models. Currently, 7.7 hours of dance data with 30 fps are available in FineDance. During the testing phase, we follow Lodge's experimental setup, generating long-sequence dances for 20 music pieces from the FineDance test set, using the first 1024 frames of motion for evaluation.

\noindent\textbf{Implementation details.} The frame number $n$ is set as 128 for efficient training. And the frame number $L$ depends on the duration of the input music. 
For the VQ-VAE, the number of Choreographic Memory Codebook, $K$ is 1024, the channel dimension of $\Tilde{z}_i$, $C_z$ is also 1024.
The temporal down-sampling rate $r$ is 4.
The motion sequences are cropped to 128 frames for training.
We use AdamW~\cite{loshchilov2017decoupled} optimizer with [${\beta}_1$, ${\beta}_2$] = [0.9, 0.99], and exponential moving constant of 0.99.  The hyper-parameter $\beta$ is set to 1. We train the VQ-VAE with a initial learning rate of 2e-4, and then reduce learning rate to 1e-5.
For the Dance GPT, We employ 18 Transformer~\cite{transformer} layers, each having a dimension of 1,024 and 16 heads. The maximum length of the code index sequence is set to 128, corresponding to approximately 17 seconds of motion. We add an `End' token to terminate prediction. Dance GPT is optimized using AdamW~\cite{loshchilov2017decoupled} with ${\beta}_1 = 0.5$ and ${\beta}_2 = 0.99$. The learning rate is initially set to 1e-4  and then decayed to 5e-6.
For the PDDM, the diffusion time steps are configured to be 50. We employ Adan~\cite{adan} as the optimizer and use the Exponential Moving Average (EMA)~\cite{ema} technique to enhance the stability of loss convergence. The learning rate is established at 1e-4.

\input{tabs/compare_sotas}

\subsection{Evaluation Metrics}
We evaluate our algorithm comprehensively from five perspectives: Motion Quality, Motion Diversity, Beat Align Score, Run Time, and the ratio of wins in user study.

\noindent\textbf{Motion Quality} primarily measures the quality of dance movements, analyzing their natural smoothness and physical realism. Motion Quality includes $\mathrm{FID}_k$, $\mathrm{FID}_g$, and Foot Skating Ratio (FSR). $\mathrm{FID}_k$ and $\mathrm{FID}_g$ are the Fréchet Inception Distances (FID) of generated dance and all dance in the dataset. The subscripts $k$ and $g$ represent the FID distances calculated using kinematic features and geometry features, respectively, and both kinematic and geometry features are extracted using the Fairmotion-tools~\cite{gopinath2020fairmotion}.
Foot Skating Ratio (FSR) is used to measure the level of contact between the feet and the ground, which measures the proportion of frames in which either foot skids more than a certain distance while maintaining contact with the ground (foot height\textless5 cm).
We follow ProHMR~\cite{zhang2023probabilistic} to calculate the ratio of penetrated vertices to the total number of vertices, obtaining the Penetration Ratio to quantitatively evaluate the character self-penetration degree.

\noindent\textbf{Motion Diversity} primarily focuses on the richness and diversity of movements, including $\mathrm{DIV}_k$ and $\mathrm{DIV}_g$. We follow the methods used in AIST++~\cite{aist++} and Bailando~\cite{bailando} to calculate the average feature distances of the generated dances to obtain DIV. The subscripts $k$ and 
$g$ represent the calculations using kinematic features and geometry features, respectively.

\noindent\textbf{Beat Align Score (BAS)}~\cite{aist++} measures the degree of rhythmic alignment between the music and dance. It evaluates how well the timing of dance movements corresponds to the beats of the music, providing insights into the synchronization of the two modalities.

\noindent\textbf{Run time} measures the efficiency of long-sequence dance generation, and we calculate the average time required to generate each dance sequence in the test set, expressed in seconds.

\noindent\textbf{Wins} is a subjective test where we ask different participants, including those with and without a dance background, to watch two sets of videos: one set containing the results generated by our method and the other set consisting of randomly selected ground truth (GT) or other state-of-the-art (SOTA) methods. Participants are then asked to choose which dance segment they believe has better quality.

\subsection{Comparison to Existing Methods}
As shown in Table~\ref{tab:results}, we compare our method with advanced music-driven dance generation algorithms. FACT is a classic autoregressive generation algorithm, while MNET builds upon FACT by incorporating a multi-genre discriminator. Bailando is an outstanding dance generation network based on VQ-VAE and GPT, demonstrating excellent generation performance. EDGE is a diffusion-based dance generation method that supports dance editing operations such as inbetweening, continuation, and joint-conditioned generation, and it has shown high quality in short-sequence dance generation.
Compared to these methods, Lodge has made significant improvements in motion quality, beat alignment, and run time.
Lodge achieved an FID of 50.00 and a Foot Skating Ratio of 2.76$\%$, representing improvements of 44.34 (47$\%$) and 16 (85.28$\%$), respectively, compared to the previous state-of-the-art methods.
$\mathrm{FID}_k$ mainly reflects the physical quality of the movements, while $\mathrm{FID}_g$ evaluates the global choreography rules' rationality. As shown, Lodge’s $\mathrm{FID}_g$ is 35.52, which, although significantly better than FACT, MNET, and EDGE, falls short of Bailando’s 28.17. 
We argue this is due to Lodge using Global Diffusion to learn overly sparse Characteristic Dance Primitives and relying on handcrafted choreography rules to enhance these Primitives, leading to weaker generalization of the learned choreography rules. 

Therefore, we draw inspiration from Bailando's choreography model, adopting the Choreography Memory Book and sequence models to learn vivid choreography patterns from data.  
Additionally, we incorporate the PDDM model and utilize the Foot Refine Block, Multi-Genre Discriminator, and Penetration Guidance, further enhancing motion quality. Ultimately, both $\mathrm{FID}_k$ and $\mathrm{FID}_g$ of Lodge++ achieve significant improvements, reaching 40.77 and 30.79, respectively. 
The Ground Truth penetration ratio is 0.6954$\%$, as the current dance dataset is collected from dancers with varying heights and body proportions. However, all dancer motions are retargeted to a standard human model for training, which introduces errors during the retargeting process and results in some penetration instances.
Due to motion freezing issues, FACT and MNET exhibit relatively low penetration ratios. In contrast, Bailando, EDGE, and Lodge show higher penetration ratios due to neglect of the hands and human vertices. 
However, humans are highly sensitive to
human motion, and even minor penetration can significantly diminish perceived realism. 
Lodge++ effectively addresses this problem with the training-free Penetration Guidance strategy, extremely reducing the penetration ratio. 
Based on the dance primitives and the PDDM, Lodge++’s BAS also achieved the best result of 0.2423.

\subsection{Ablation Study}
\noindent\textbf{The Global Choreography Network:}
The Global Choreography Network (GCN) can efficiently model long-term choreography patterns. On one hand, it ensures that the generated dances is better choreography structure; on the other hand, it guides the local diffusion model to generate more expressive dance movements.
We compared the method without GCN (i.e., not using the global choreography network and instead generating long-sequence dances with an autoregressive approach) to the method that uses the GCN to generate Dance Primitives, which then guides the diffusion model to produce high-quality dances. As shown in Table ~\ref{tab:GCN}, it is evident that GCN significantly improves motion quality, motion diversity, and beat alignment.
\input{tabs/ablation_GCN}
\input{tabs/ablation_PDDM}

\noindent\textbf{The PDDM:}
As shown in Table ~\ref{tab:GCN}, w.o GCN refers to the method without the global choreography network, where long-sequence dances are generated using diffusion inpainting and an autoregressive approach, resulting in poor motion quality, motion diversity, and beat alignment. In contrast, w. GCN refers to the incorporation of the global choreography network, which effectively models choreography rules and guides PDDM to generate dances with better motion quality, greater diversity, and stronger beat alignment.

% w. GCN incorporates the global choreography network, using the Dance Primitives it generates to guide PDDM in producing long-sequence dances that adhere to choreography rules. Significant improvements in $\mathrm{FID_k}$, $\mathrm{DIV_k}$, and $\mathrm{BAS}$ can be observed.

\noindent\textbf{The Denoise Network of PDDM:} To further improve motion quality and ensure consistency in dance style, we incorporated a Foot Refine Block and a Multi-Genre Discriminator. As shown in Table 4, the inclusion of the Foot Refine Block reduced the $\mathrm{FID}_k$ from 52.23 to 43.72, and the $\mathrm{FSR}$ dropped from 5.92 to 3.31. When both the Foot Refine Block and the Multi-Genre Discriminator were applied, the overall best results were achieved with $\mathrm{FID}_k$ reduced to 40.77, $\mathrm{FSR}$ remaining at 4.11$\%$, and the BAS reaching 0.2423.
\input{tabs/ablation_DenoiseNet}

\noindent\textbf{The Peneration Guidance:}
The purpose of Penetration Guidance is to reduce self-penetration between the hands and body. 
As shown in table~\ref{tab:peneguidance}, Penetration Guidance significantly reduces the Penetration Rate (PR) from 0.1928$\%$ to 0.0079$\%$, while maintaining $\mathrm{FID}_k$ and improving $\mathrm{BAS}$. 
\input{tabs/ablation_PeneGuide}

% Lodge adopts a two-stage coarse-to-fine diffusion framework.
% : first, Global Diffusion generates Characteristic Dance Primitives, which are then enhanced based on choreography principles to guide Local Diffusion in generating long-sequence dances in parallel. Lodge has made breakthroughs in long-sequence dance generation, but its reliance on predefined choreography rules limits its adaptability to different dance genres. 

\begin{figure*}[t!]
    \centering
    \includegraphics[width=\textwidth]{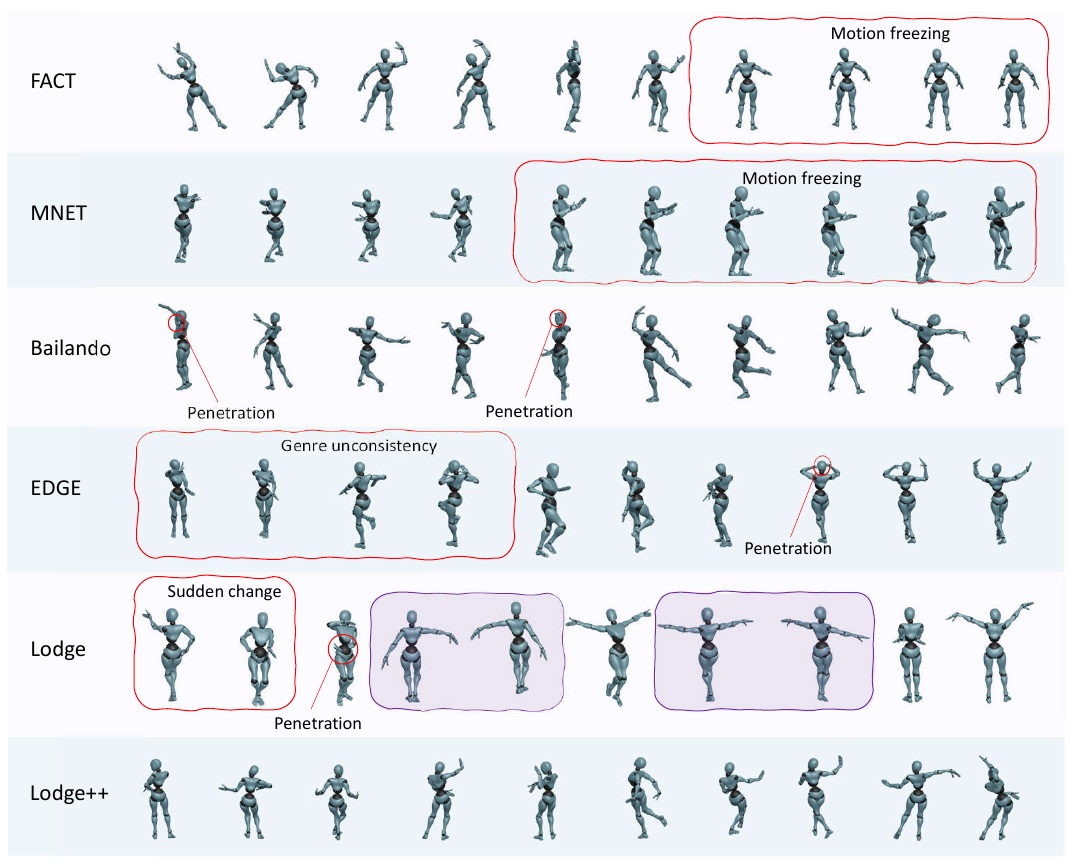}
    \caption{Visual comparison of dance sequences generated by different algorithms. The red boxes highlight various motion issues, such as motion freezing and sudden changes in movements. The purple boxes indicate that Lodge generates dances based on predefined symmetrical choreography rules. While this approach ensures a certain level of choreographic coherence, it also hinders the network from learning more complex and vivid choreography patterns. Lodge++ demonstrates a professional level of choreography while maintaining high-quality local motion details.}%\vspace{-3mm}
    \label{fig:compare}
    % \vspace{-1.5em}
\end{figure*}

\subsection{Qualitative analysis}
\subsubsection{User study} We conducted a user study to qualitatively compare the generated results of different algorithms. We invited 40 participants, including individuals with no dance background, those with basic dance experience, and professional dancers. They were asked to watch 17 pairs of dance videos. In each pair, one video was generated by Lodge++, and the other was either the ground truth or a result produced by other algorithms. The order of the videos was randomized to ensure a fair comparison.
Participants were instructed to judge which dance motion had better overall quality. The experimental results are recorded in the  ``Lodge++ wins" column of Table~\ref{tab:results}. Due to obvious motion quality issues in other algorithms—such as motion freezing, sudden changes, foot-ground contact problems, and penetrations 
Lodge++ significantly outperformed other methods in most evaluations, winning against EDGE in 84.3$\%$ of comparisons and against Bailando in 74.5$\%$ of comparisons.
Additionally, Lodge++ outperformed Lodge in 65.1$\%$ of the cases. This is mainly because Lodge++ can learn more complex and vivid choreography patterns, alleviating issues like penetration and sudden motion changes.

\subsubsection{Visual Analysis with Existing Methods} 
\begin{figure}[t]
\begin{center}
\includegraphics[width=0.45\textwidth]{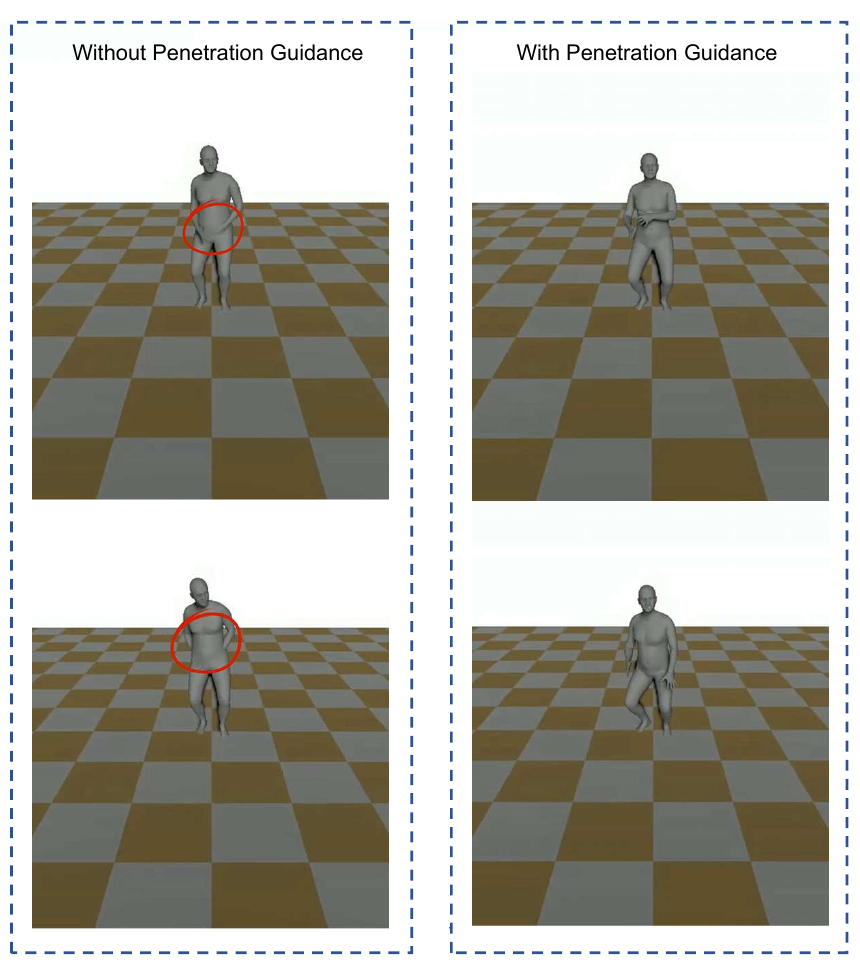}
% \vspace{-3mm}
\caption{A visual ablation study of the Penetration Guidance. When the Primitives-based Dance Diffusion does not use Penetration Guidance, hand-body penetration easily occurs, as shown in the red circle. In contrast, Penetration Guidance effectively resolves this issue.}	
\label{fig:peneguidance}
\end{center}
\end{figure}
As shown in Fig.~\ref{fig:compare}, we conducted a visual analysis of our algorithm's performance compared to other algorithms. More visual analyses can also be found in the supplementary video.
FACT~\cite{aist++} and MNET~\cite{kim2022brand}, which adopt an autoregressive framework to generate dance frame by frame, experience motion freezing issues after a few seconds. Bailando~\cite{bailando} is a choreography network based on VQ-VAE and GPT that can learn complex choreography patterns. However, its pre-trained VQ-VAE makes it difficult for the subsequent GPT network to optimize fine-grained motion quality, resulting in penetration and foot-ground contact issues. 
EDGE~\cite{edge} is a diffusion-based dance generation network that creates long dance sequences by generating multiple segments in parallel and seamlessly stitching them together.
However, EDGE tends to have sudden motion changes at the splicing frames. 
As shown in Fig.~\ref{fig:compare}, EDGE has difficulty ensuring overall style consistency across different generated segments and also suffers from penetration problems.

Lodge primarily uses characteristic dance primitives and diffusion guidance to generate expressive long-sequence dances that conform to choreography rules. However, Fid.~\ref{fig:compare} shows that Lodge sometimes produces sudden motion changes under the guidance of dance primitives and also has penetration issues. The purple box in Fid.~\ref{fig:compare} illustrates that Lodge can generate dances with symmetrical structures based on basic choreography rules. While this endows Lodge with fundamental choreographic rationality, it also hinders it from learning more complex and vivid choreography patterns. As observed, Lodge++ effectively addresses the aforementioned issues, exhibiting complex choreography patterns suitable for various dance styles while ensuring the quality of local motions.

\subsubsection{Visual Analysis of the Penetration Guidance} 
As shown in Figure~\ref{fig:peneguidance}, we rendered a comparison of the effects with and without Penetration Guidance. Since we calculate the penetration ratio metric and optimize movements based on the SMPLX model, we use the rendering of the original SMPLX model to showcase the comparison. 
Without penetration guidance, hand movements frequently penetrate the body. 
The proposed Penetration Guidance effectively resolves self-penetration by fine-tuning arm angles based on gradients computed from the SDF.

%% file: tabs/compare_sotas.tex
\begin{table*}
\caption{Compare with the state-of-the-art methods on the FineDance dataset. Wins is the ratio of victories Lodge++ achieved in the user study.}
\resizebox{\linewidth}{!}
{
\begin{tabular}{lcccccccccc}
    \toprule[1.5pt]
    \noalign{\smallskip}
    \multirow{2}*{Method} & \multicolumn{4}{c}{Motion Quality} & \multicolumn{2}{c}{Motion Diversity} & \multirow{2}*{BAS$\uparrow$} & \multirow{2}*{Run Time$\downarrow$} & \multirow{2}*{Wins$\uparrow$} \\ \cmidrule(lr){2-5} \cmidrule(lr){6-7}
    & $\mathrm{FID}_k\downarrow$ & $\mathrm{FID}_g\downarrow$ & Foot Skating Ratio$\downarrow$ & Penetration Ratio$\downarrow$ & $\mathrm{Div}_k\uparrow$ & $\mathrm{Div}_g\uparrow$ & & & \\
    \noalign{\smallskip}\midrule
    \noalign{\smallskip}
        Ground Truth & / & / & 6.22$\%$ & 0.6954$\%$& 9.73 & 7.44 & 0.2120 & / & 45.6$\%$ \\
    \midrule
    FACT\cite{aist++} & 113.38 & 97.05 & 28.44$\%$ & 0.1001$\%$ & 3.36 & 6.37 & 0.1831 & 35.88s & 95.6$\%$ \\
    MNET\cite{kim2022brand} & 104.71 & 90.31 & 39.36$\%$ & 0.2796$\%$ & 3.12 & 6.14 & 0.1864 & 38.91s & 94.4$\%$ \\
    Bailando\cite{bailando} & 82.81 & \textbf{28.17} & 18.76$\%$ & 0.8032$\%$& 7.74 & 6.25 & 0.2029 & 5.46s & 74.5$\%$ \\
    EDGE\cite{edge} & 94.34 & 50.38 & 20.04$\%$ & 0.9467$\%$ & \textbf{8.13} & \textbf{6.45} & 0.2116 & 8.59s & 84.3$\%$ \\
    \midrule
    \textbf{Lodge}~\cite{li2024lodge} & 50.00 & 35.52 & \textbf{2.76$\%$} & 0.5591$\%$ & 5.67 & 4.96 & 0.2269 & \textbf{4.57}s & 65.1$\%$ \\
    % \textbf{Lodge} (DDPM) & 45.56 & 32.29 & {5.01$\%$} & 6.75 & 5.64 & \textbf{0.2397} & 30.93s & xx.x$\%$ \\
    \textbf{Lodge++} & \textbf{40.77} & 30.79 & {4.11$\%$} & \textbf{0.0079}$\%$& 5.53 & 5.01 & \textbf{0.2423} & 6.24s & / \\
    \bottomrule[1.5pt]
\end{tabular}
}
\label{tab:results}
\end{table*}

%% file: tabs/ablation_GCN.tex
\begin{table}[!htbp]
\caption{Ablation study of the Global Choreography Network.}
% \vspace{-2mm}
\centering
\resizebox{0.4\textwidth}{!}{
	\begin{tabular}{lcccccccc}
		\toprule [1pt] 
		Method &   $\mathrm{FID}_k\downarrow$ &   $\mathrm{Div}_k\uparrow$  &  BAS $\uparrow$ \\
		\noalign{\smallskip}\hline\noalign{\smallskip}
            Ground Truth  &/  &9.73& 0.2120\\
            \noalign{\smallskip}\hline\noalign{\smallskip}
            w.o GCN    &63.38  &4.50 & 0.2156\\ 
            w. GCN   &40.77  &5.53  & 0.2423\\
		\bottomrule [1pt] 
               \noalign{\smallskip}
	\end{tabular}
	}
        % \vspace{-2mm}
	\label{tab:GCN}   
        % \vspace{-2mm}
\end{table}

%% file: tabs/ablation_PDDM.tex
\begin{table}[!htbp]
\caption{Ablation study of the PDDM.}
% \vspace{-2mm}
\centering
\resizebox{0.4\textwidth}{!}{
	\begin{tabular}{lcccccccc}
		\toprule [1pt] 
		Method &   $\mathrm{FID}_k\downarrow$ &   $\mathrm{Div}_k\uparrow$  &  BAS $\uparrow$ \\
		\noalign{\smallskip}\hline\noalign{\smallskip}
            Ground Truth  &/  &9.73& 0.2120\\
            \noalign{\smallskip}\hline\noalign{\smallskip}
            Diffusion Guidance    &53.48  &4.95 & 0.2071\\ 
            PDDM   &40.77  &5.53  & 0.2423\\  
		\bottomrule [1pt] 
               \noalign{\smallskip}
	\end{tabular}
	}
        % \vspace{-2mm}
        \label{tab:softguidance}   
        % \vspace{-2mm}
\end{table}

%% file: tabs/ablation_DenoiseNet.tex
\begin{table}[!htbp]
\caption{Ablation study of the Denoise Network of PDDM.  'F' means the Foot Refine Block, 'G' means Multi-Genre Discriminator. 'FSR' means the Foot Skating Rate.}
% \vspace{-2mm}
\centering
\resizebox{0.4\textwidth}{!}{
	\begin{tabular}{lcccccccc}
	\toprule [1pt] 
	\multicolumn{2}{c}{Ablations} & \multicolumn{4}{c}{Metrics} \\ \cmidrule(lr){1-2}   \cmidrule(lr){3-6}   
        F & G &   $\mathrm{FID}_k\downarrow$ &   $\mathrm{Div}_k\uparrow$  &  FSR $\downarrow$ &  BAS $\uparrow$ \\
	\noalign{\smallskip}\hline\noalign{\smallskip}
         &   &52.23  &4.53 &5.92$\%$ &0.2306\\
         \checkmark &    &43.72  &5.15  &3.31$\%$ &0.2334\\
        &  \checkmark &45.50  &4.89 &5.99$\%$ &0.2313\\
		\checkmark & \checkmark & 40.77  &5.53  &4.11$\%$ &0.2423\\
		\bottomrule [1pt] 
               \noalign{\smallskip}
	\end{tabular}
	}
        % \vspace{-2mm}
        \label{tab:Primitives}   
    % \vspace{-2mm}
\end{table}

%% file: tabs/ablation_PeneGuide.tex
\begin{table}[!htbp]
\caption{Ablation study of the Penetration Guidance(PG). 'PR' means the Penetration Rate.}
% \vspace{-2mm}
\centering
\resizebox{0.4\textwidth}{!}{
	\begin{tabular}{lcccccccc}
		\toprule [1pt] 
		Method &   $\mathrm{FID}_k\downarrow$ &   $\mathrm{Div}_k\uparrow$ & PR &  BAS $\uparrow$ \\
		\noalign{\smallskip}\hline\noalign{\smallskip}
            Ground Truth  &/  &9.73& 0.6954$\%$ & 0.2120\\
            \noalign{\smallskip}\hline\noalign{\smallskip}
            w.o PG    &38.92  &5.51 & 0.1928$\%$  & 0.2370\\ 
            w. PG   &40.77  &5.53 & 0.0079$\%$  & 0.2423\\  
		\bottomrule [1pt] 
               \noalign{\smallskip}
	\end{tabular}
	}
        % \vspace{-2mm}
    \label{tab:peneguidance}   
        % \vspace{-2mm}
\end{table}

%% file: sec/5_Conclusion.tex
\section{Conclusion}
% Current music-driven 3D dance generation algorithms struggle to balance computational efficiency, global choreography patterns, and local movement quality. To address these challenges, we proposed Lodge, which introduced a two-stage coarse-to-fine choreography framework and characteristic dance primitives as an intermediate-level representation. Building upon this, we further developed Lodge++, enhancing the framework to better capture complex choreography patterns and improve the physical realism of the generated dances. 
Current music-driven 3D dance generation algorithms often struggle to balance computational efficiency, global choreography, and local movement quality. To address these issues, we introduced Lodge, a two-stage coarse-to-fine framework that incorporates dance primitives as intermediate representations. Building on this, Lodge++ further enhances choreography complexity and the physical realism of the generated dances.

In Lodge++, we utilize a VQ-VAE and GPT based Global Choreography Network to learn the coarse-grained relationships between the whole music and dance, generating coarse-grained dance.
By extracting $\bm{d_h}$ and $\bm{d_s}$ from the generated coarse-grained dance, we obtain the dance primitives.
These dance primitives effectively convey complex and vivid global choreography patterns, making them applicable to various dance genres.
Furthermore, we introduce the Primitive-based Dance Diffusion Model, which generates high-quality, long-sequence dances in parallel. By denoising from noise that includes dance primitives, this model simplifies the denoising process and reduces the gap between training and inference, effectively addressing the motion abruptness issues encountered in the previous Lodge method. Additionally, we present the Foot Refine Block to enhance foot-ground contact quality, a Multi-Genre Discriminator to maintain genre consistency, and a novel SDF-based Penetration Guidance module to alleviate self-penetration issues between the hands and body.

Extensive qualitative and quantitative experiments validate the effectiveness of our method. The long-sequence dances generated by Lodge++ achieved $\mathrm{FID}_k$ (fine-grained movement quality) and $\mathrm{FID}_g$ (global choreography quality) scores of 40.77 and 30.79, respectively. Notably, the $\mathrm{FID}_k$ score of 40.77 represents the best result in fine-grained movement quality, while the $\mathrm{FID}_g$ score of 30.79 is only 2.62 points (9.3$\%$) higher than Bailando’s 28.17, demonstrating that our model maintains excellent global choreography while achieving optimal fine-grained movement quality.
We observe that Bailando effectively learns complex choreography patterns, but its movement quality metrics, such as $\mathrm{FID}_k$, Foot Skating Ratio, and Penetration Rate, are relatively low. Our approach leverages Bailando’s strengths by using a Global Choreography Network to capture complex and vivid choreography patterns. Additionally, Lodge++’s proposed Foot Refine Block, Multi-Genre Discriminator, and Penetration Guidance significantly improve performance on metrics such as Foot Slide Ratio (FSR) and Penetration Rate (PR). User studies further indicate that the quality of dances generated by our algorithm surpasses those produced by existing methods.

Despite these advancements, our method has limitations. It currently lacks fine-grained facial expressions and finger movements, which are important components of dance performance. Additionally, our method focuses on body movement data and does not incorporate elements such as dancers' costumes and props. Future work could explore integrating detailed facial and hand movements and extending the framework to include visual appearance aspects like costumes and props, aiming to create more immersive and comprehensive dance generation systems.

%% file: sec/6_Appendix.tex
% or


% \appendices

% \section{}

% \ifCLASSOPTIONcompsoc
%   % The Computer Society usually uses the plural form
%   \section*{Acknowledgments}
% \else
%   % regular IEEE prefers the singular form
% \fi

% This work was supported in part by the Peng Cheng Laboratory (PCL2023A10-2), in part by the Natural Science Foundation of China (No. 62377004), in part by the National Natural Science Foundation of China (No. 62306165), in part by the Fundamental Research Funds for the Central Universities (No. 2233100028).

\ifCLASSOPTIONcaptionsoff
  \newpage
\fi